%% file: main.tex
\documentclass{article} 
\usepackage{iclr2026_conference,times}

\input{math_commands.tex}

\definecolor{refblue}{RGB}{0,51,153}
\PassOptionsToPackage{hidelinks}{hyperref}
\PassOptionsToPackage{colorlinks=true,citecolor=refblue,linkcolor=black,urlcolor=blue}{hyperref}
\usepackage{hyperref}
\usepackage{url}
\usepackage{graphicx}
\usepackage{booktabs}
\usepackage{multirow}
\usepackage{xcolor}
\usepackage[table]{xcolor}
\usepackage{amssymb}
\usepackage{pifont}
\usepackage{wrapfig}
\usepackage{arydshln}
\usepackage{graphicx}
\usepackage{wrapfig}
\usepackage{lipsum} 
\usepackage{tabularx}

\definecolor{myblue}{RGB}{187,225,245}
\definecolor{lightgreen}{RGB}{229,235,227}
\definecolor{darkgreen}{RGB}{199,213,195}
\title{ODI-Bench: Can MLLMs Understand Immersive Omnidirectional Environments?}

\author{Liu Yang$^{1, }$\thanks{Equal contribution.}, Huiyu Duan$^{1, *,} \thanks{Corresponding authors.}$, Ran Tao$^{2}$, Juntao Cheng$^{1}$, Sijing Wu$^{1}$, Yunhao Li$^{1}$, Jing Liu$^{3}$, \\ \textbf{Xiongkuo Min$^{1}$}, \textbf{Guangtao Zhai$^{1}$}\\
$^{1}$Shanghai Jiao Tong University \\
$^{2}$Xinjiang University\\
$^{3}$Tianjin University
}

\iclrfinalcopy 
\begin{document}

\maketitle
\input{figures/teaser}
\vspace{-6pt}
\begin{abstract}
\vspace{-6pt}
Omnidirectional images (ODIs) provide full 360$^{\circ} \times$ 180$^{\circ}$ view which are widely adopted in VR, AR and embodied intelligence applications. While multi-modal large language models (MLLMs) have demonstrated remarkable performance on conventional 2D image and video understanding benchmarks, their ability to comprehend the immersive environments captured by ODIs remains largely unexplored. To address this gap, we first present \textbf{{ODI-Bench}}, a novel comprehensive benchmark specifically designed for omnidirectional image understanding. ODI-Bench contains \textit{2,000} high-quality omnidirectional images and over \textit{4,000} manually annotated question-answering (QA) pairs across \textit{10} fine-grained tasks, covering both general-level and spatial-level ODI understanding. Extensive experiments are conducted to benchmark \textit{20} representative MLLMs, including proprietary and open-source models, under both \textit{close-ended} and \textit{open-ended} settings. Experimental results reveal that current MLLMs still struggle to capture the immersive context provided by ODIs. To this end, we further introduce \textbf{Omni-CoT}, a training-free method which significantly enhances MLLMs’ comprehension ability in the \underline{omni}directional environment through \underline{c}hain-\underline{o}f-\underline{t}hought reasoning across both textual information and visual cues. Both the benchmark and the code will be released at \textcolor{refblue}{https://github.com/IntMeGroup/ODI-Bench}.
\end{abstract}
\vspace{-2em}
\section{Introduction}
\vspace{-6pt}

\input{tabs/table2-comparison}
360$^{\circ}$ omnidirectional images (ODIs) have gained increasing attention nowadays. Unlike conventional 2D images with limited field of views (FoVs), ODIs provide a full 180$^{\circ} \times$ 360$^{\circ}$ FoV with rich scene information, enabling fully immersive viewing. Thus, ODIs are widely used in virtual reality (VR), augumented reality (AR), spatial navigation, and hold great potential for embodied intelligence~\citep{yang2025omni,zheng2025panorama}. Although recent advances in multi-modal large language models (MLLMs) have led to significant progress in conventional image understanding across various benchmarks~\citep{liu2024mmbench,yu2023mm}, their ability to comprehend ODIs has not been comprehensively evaluated, with existing benchmarks remaining insufficient.
Compared to conventional images, ODIs capture substantially richer visual information from omnidirectional scenes and require higher-level spatial reasoning in immersive environments beyond a single front-view perspective. These unique characteristics make ODI understanding a distinct and difficult research challenge, highlighting the necessity to systematically evaluate MLLMs on this task.

Though a limited number of ODI understanding benchmarks have been proposed as shown in Table~\ref{tab2:comparison}, they generally suffer from one or more of the following issues: (1) \textbf{Low resolution}: since many applications (such as VR, autonomous driving) requires high-resolution ODIs to provide immersive viewing experience or 360$^{\circ}$ details, benchmarks with low resolution are impractical for real-world application~\citep{chou2020visual,dongfang2025multimodal}; (2) \textbf{Limited scene diversity}: some benchmarks are developed with the assistance of 3D-annotated indoor datasets~\citep{chou2020visual,dongfang2025multimodal}, focousing only on indoor environments, or even unrealistic synthetic indoor scenes with blurry top-bottom views; (3) \textbf{Constrained question domains}: Existing benchmarks are automatically annotated, either leveraging existing 3D datasets or curated pipelines for question generation, thus tend to exhibit strong textual biases and provide relatively narrow or simplistic question types~\citep{zhou2025dense360}; (4) \textbf{Viewpoint limitation}: for spatial understanding, existing ODI benchmarks are primarily designed from an egocentric perspective, neglecting allocentric viewpoints and the simulation of user interactions. As a result, they fall short in evaluating the embodied aspects of spatial understanding, which are critical for advancing embodied intelligence and interactive multimodal systems.

To address these gaps, we introduce \textbf{ODI-Bench}, a novel ODI-oriented benchmark designed to comprehensively evaluate both the general-level and spatial-level understanding capabilities of MLLMs. The question–answer pairs are derived from two complementary sources: (1) a rigorously designed automated pipeline that generates reliable instance-level QA pairs, which are further checked and refined by human experts, and (2) high-quality human annotations produced by three domain experts, whose works are carefully cross-checked to ensure reliability. The final benchmark contains \textit{2,000} high quality real-life omnidirectional images, covering diverse indoor and outdoor scenes. \textit{10} representative tasks are proposed to facilitate fine-grained and multi-perspective evaluation of the performance of MLLMs under ODI settings, partially illustrated in Figure~\ref{figs:teaser}.

Unlike previous benchmarks that restrict the evaluation of each task to either a close-ended (multiple-choice or true/false) or an open-ended QA setting, \textbf{ODI-Bench} evaluates every task under \textit{both} settings. This dual-format design enables a comprehensive and comparative assessment, capturing both the recognition accuracy under constrained choice conditions and the model’s generative reasoning ability in unconstrained scenarios. Experimental results demonstrate that MLLMs still struggle to comprehend the immersive environment presented by ODIs. To this end, we further propose a training-free chain-of-thought framework, termed \textbf{Omni-CoT}, to improve MLLMs' understanding capabilities on ODIs through step-by-step reasoning with viewpoint-guided scene interpretation and visual cue based refinement. This approach significantly enhances MLLMs’ comprehension on ODIs across both general and spatial-level tasks. Our contributions are summarized as follows:
\vspace{-6pt}
\begin{itemize}
    \item We introduce \textbf{ODI-Bench}, a comprehensive benchmark for evaluating MLLMs on omnidirectional image understanding, which consists of 2,000 high-quality omnidirectional images and over 4,000 QA pairs across 10 fine-grained tasks, covering both general and spatial-level ODI understanding.
    \item We conduct an in-depth study to evaluate the ODI comprehension ability of 20 leading MLLMs on our ODI-Bench, using both close-ended and open-ended settings to make comprehensive and comparative analysis. Experimental results reveal the challenges of MLLMs in understanding immersive ODI scenes.
    \item We propose \textbf{Omni-CoT}, a training free strategy to enhance MLLMs' comprehension capabilities on omnidirectional scenes through chain-of-thought reasoning. Experimental results demonstrate the effectiveness of the proposed framework on both propritary and open-sourced models.
\end{itemize}
\vspace{-9pt}
\section{Related Works}
\vspace{-6pt}
\subsection{General Understanding Benchmarks}
\vspace{-6pt}
With the advancement of MLLMs, there is an increasing need for comprehensive and systematic evaluation of their visual understanding capabilities. A number of benchmarks have been developed to assess the general-level comprehension ability of MLLMs~\citep{liu2024mmbench,yu2023mm, duan2025finevq}. However, as the performance of MLLMs has significantly improved, such general benchmarks are no longer sufficient for thorough ability assessment. More recently, new benchmarks are proposed to evaluate the spatial understanding ability of MLLMs~\citep{yang2025thinking,liu2025ssr}, presenting new challenges for MLLMs on spatial understanding tasks. However, while most of these benchmarks focus on 2D images or NFoV videos, the benchmarks specifically designed for omnidirectional images are still scarce. Given their unique format and application scenarios, the ability to understand ODIs holds great potential for advancing not only MLLMs but also vision-language-action (VLA) models.
\vspace{-6pt}
\subsection{Omnidireectional Image Understanding Benchmarks}
\vspace{-6pt}
A limited number of ODI understanding benchmarks are proposed to evaluate MLLMs' understanding of this unique type of images. Dense360-Bench~\citep{zhou2025dense360} introduces a QA curation pipeline and further constructs a benchmark for general-level grounding and captioning tasks on ODIs, but such tasks remain superficial and fall short in adequately evaluating spatial understanding abilities. VQA 360$^{\circ}$~\citep{chou2020visual} constructs a benchmark for simple ODI understanding tasks, but the image resolution is too low (1024 $\times$ 512), constraining its applicability in real-world scenarios. OSR-Bench~\citep{dongfang2025multimodal} develops a pipeline for generating ODI spatial comprehension QA pairs from 3D datasets, yet it focuses solely on synthetic low-resolution indoor scenes, limiting its applicability. In contrast, our ODI-Bench is the first to comprehensively benchmark both the general-level and spatial understanding capabilities of MLLMs on ODIs, with carefully manually curated QA-pairs assisted by an automatic annotation pipeline, encompassing both high-quality indoor and outdoor scenes.
\vspace{-6pt}
\section{ODI-Bench}\label{sec:bench_process}
\vspace{-3pt}
\subsection{Overview of ODI-Bench}
In this section, we introduce \textbf{ODI-Bench}, a benchmark for comprehensive evaluation of MLLMs on omnidirectional image understanding.ODI-Bench consists of \textit{2,000} real-world omnidirectional images covering diverse indoor and outdoor scenes, along with \textit{4,254} question-answering pairs across \textit{10} fine-grained tasks, offering both general-level and spatial-level ODI understanding evaluation.
\vspace{-6pt}
\subsection{Image Collection}
\vspace{-3pt}
The images in our benchmark are primarily web-crawled from Flickr and carefully selected to ensure both quality and diversity. The distribution of images is presented in Figure  \ref{figs:distribution} (a) and (b). Compared with existing omnidirectional image understanding benchmarks \citep{dongfang2025multimodal, chou2020visual} which are largely restricted to indoor environments and predominantly depict house-like scenes rendered from 3D datasets, our benchmark covers diverse indoor and outdoor scenes, ranging from human activity to natural landscapes, hereby enabling a comprehensive evaluation across diverse scenarios. In addition, some exsiting benchmarks suffer from low image quality. For instance, VQA 360$^{\circ}$~\citep{chou2020visual} consists of ODIs with blurry top and bottom views and the image resolutions are limted to 1K, which restricts their practical value in real-world applications. In contrast, as presented in Figure \ref{figs:distribution} (b), our benchmark is high quality, with sufficient resolution to ensure both practical applicability and reliable benchmarking.
\vspace{-6pt}
\subsection{Fine-grained Task Defination}

\input{figures/distribution}
\vspace{-3pt}
Unlike previous works that evaluate MLLMs’ comprehension abilities on ODIs either at the general level~\citep{zhou2025dense360} or the spatial level~\citep{dongfang2025multimodal}, or through a simple combination of the two~\citep{chou2020visual}, we carefully design 10 fine-grained tasks tailored for comprehensive ODI understanding, covering both general-level and spatial-level aspects, as shown in Figure \ref{figs:distribution}.
\vspace{-3pt}
\paragraph{General-level ODI understanding.}
\vspace{-3pt}
Inspired by conventional 2D image understanding tasks, we propose five main general-level tasks to evaluate MLLMs’ comprehension on common ODI scenarios. These tasks typically impose relatively low spatial reasoning requirements, while the main challenges arise from the massive amount of visual information and distorted projections inherent in ODIs. Among them, we define instance-level tasks, \textit{i.e., }\textit{object-attribute} and \textit{human-attribute}, to assess the models’ ability to accurately localize instances and extract visual information across wide fields of view. In addition, we introduce \textit{counting} and \textit{existence} tasks to measure the models’ global perception capabilities on omnidirectional images. Finally, we define a omnidirectional \textit{OCR} task to evaluate the models’ capability to extract textual information under distorted perspectives and high-resolution conditions, including cross-view scenarios.
\vspace{-3pt}
\paragraph{Spatial-level ODI understanding.}
\vspace{-3pt}
Omnidirectional images project immersive 3D scenes onto a 2D plane, leading to substantial differences in spatial perception compared to conventional 2D images. In 2D images, the viewer is constrained to a single front-facing perspective, while ODIs provide a full 360$^{\circ}$ field of view encompassing front, back, left, right, top, and bottom perspectives. To evaluate the capability of MLLMs in handling such unique spatial characteristics, we design dedicated spatial-level ODI understanding tasks. These include \textit{egocentric view orientation} and \textit{relative direction} tasks, which adopt the viewer’s own perspective, as well as \textit{allocentric view orientation} and \textit{scene simulation} tasks, which involve perspective-taking from another agent or a virtual viewpoint. Furthermore, due to the equirectangular projection (ERP) format of ODIs, spatial relationships and motion trajectories often become distorted and confusing. To address this, we introduce an ODI-reasoning task, specifically designed to assess MLLMs’ ability to understand and interpret ERP-related spatial properties in ODIs, as illustrated in Figure~\ref{figs:teaser}.
\subsection{Question-answering Annotation}
\vspace{-6pt}
Instead of conventional phrasing in 2D images benchmarks (\textit{e.g.}, ``What is [A] on \textit{the right side of the image} doing?''), we adopt an \textit{immersive question design} tailored to the characteristics of omnidirectional images, \textit{i.e.}, ``What is [A] on \textit{my right} doing?''. This first-person phrasing not only aligns with the natural immersive viewing experience of ODIs but also serves to evaluate the ability of MLLMs to understand and utilize ODIs in interactive environments.
For the QA construction process, we adopt both automatic pipelines and manual annotation process for different tasks.

For instance-level QA generation, \textit{i.e.,} object attribute and  human attribute, an automatic pipeline is adopted as presented in Figure \ref{figs:pipeline}. The ERP-formatted image is first cubemap-projected into 6 non-overlapping viewpoints for lower distortion. After that, GroundedSAM~\citep{ren2024grounded} is adopted to segment instances in each view, producing both segmentation masks and instance labels. To ensure precise instance segmentation, instances spanning multiple views are filtered out. The remaining instances are cropped based on the segmentation masks and fed into Qwen2.5-VL-72B~\citep{bai2025qwen2} for detailed caption. To ensure reliable instance selection, only those instances whose predicted categories by GroundedSAM are consistent with the descriptions from Qwen-VL-72B are retained. These captions are further utilized by GPT-4o to generate QA pairs. All the generated QAs are manually refined to ensure (i) unique reference, so that each question clearly targets a single instance; (ii) answer accuracy, guaranteeing the correctness of the provided answers.

For more complex tasks including counting, view-orientation and relative direction, \textit{etc.}, automatic generation is not trust-worthy enough, thus manual annotation is employed for precision. The annotation process took one month in VR environments with three expert participants, whose annotations were cross-checked to guarantee accuracy.
To construct close-ended evaluation, GPT-4o is adopted to generate 3 distractor options based on the QA pairs and the corresponding omnidirectional images. Each multiple-choice question is accompanied by three distractors, which are further assessed by human annotators to ensure their plausibility and to avoid semantic overlap with the correct answer. To mitigate model bias, the options are randomly shuffled, thereby ensuring a balanced distribution of correct answers across choices A to D.
\vspace{-12pt}
\section{Experiment}
\vspace{-6pt}
\input{figures/pipeline}
\subsection{Evaluation Setup}\label{sec:bench_exp}
\vspace{-3pt}
\input{tabs/table1-main}
\input{tabs/table3-openended}
We conduct comprehensive experiments on 20 leading MLLMs with different architectures and parameter scales on our ODI-Bench. The models can be categorized into two groups: (1) proprietary models, including GPT-4o \citep{hurst2024gpt}, o3 \citep{openai2025o3systemcard}, Gemini \citep{comanici2025gemini}, \textit{etc.} (2) open-sourced models, including InternVL series~\citep{zhu2025internvl3}, Qwen-VL series~\citep{bai2025qwen2}, LLaVA-NeXT~\citep{li2024llava}, LLava-OneVision~\citep{li2024llavaonevision}, \textit{etc.} All models are evaluated using the same prompt template provided in the Appendix.

We believe that model performance may vary under different evaluation settings, \textit{i.e.,} close-ended and open-ended conditions. Unlike prior benchmarks, we benchmark all models across all tasks using both close-ended formats (multiple-choice or yes/no) and open-ended formats, providing a comprehensive and comparrative assessment of their capabilities.
For close-ended benchmark, model performances are measured by their accuracy on multi-choice or yes/no questions. For open-ended benchmark, we adopt the \textit{LLM-based evaluator}~\citep{yu2023mm}, as detailed in the Appendix E.3.
Finally, we report the average score for each task.
\vspace{-6pt}
\subsection{Main Results}
\vspace{-6pt}
Close-ended and open-ended performances of all models across all tasks are reported in Table~\ref{tab1-main} and Table~\ref{tab3-open-ended}, respectively. For the close-ended evaluation, we additionally include GPT-4o without image input (Blind GPT-4o) and chance-level accuracy as baselines for comparison. 
\vspace{-6pt}
\subsubsection{Overall Performance}
\vspace{-6pt}
As illustrated in Table \ref{tab1-main}, proprietary models achieve the strongest overall performance under both the close-ended and open-ended evaluation settings, where ChatGPT o3 attaining the top overall score of 62.62 and 49.53, respectively. Open-source models also show competitive results, in which Qwen2.5-VL-72B and InternVL3-78B even outperforming GPT-4o. However, the results are still \textbf{far from satisfactory}, revealing that current MLLMs still struggle to comprehend the immersive environments presented by omnidirectional images. Besides, the best-performing model (o3) exceeds the Blind GPT-4o baseline by less than 30\% accuracy under close-ended setting, suggesting that MLLMs still struggle to comprehend the rich visual information conveyed by ODIs.
\vspace{-6pt}
\subsubsection{Task-wise Performance}
\vspace{-6pt}
From both the close-ended and open-ended evaluations, it is evident that \textbf{ODI spatial understanding is substantially more challenging than general understanding.} For general-level tasks such as attribute recognition, existence verification, and OCR, the complexity of immersive environments increases task difficulty; nevertheless, models can still partially interpret ERP images from a 2D perspective and thereby produce correct answers, which aligns with their already strong capabilities in 2D general-level understanding.
However, for tasks more closely related to spatial comprehension, \textit{i.e.}, counting, model performance drops significantly (by about 20\% compared with attribute recognition tasks).

The challenge becomes even more pronounced for tasks that fully rely on immersive spatial comprehension. Since current MLLMs are primarily trained on 2D data, their spatial reasoning capabilities are inherently limited. These limitations become especially obvious when it comes to omnidirectional comprehension requiring immersive spatial understanding, where model performance drops greatly compared to general-level tasks, only slightly above the random choice baseline. This issue is particularly evident in non-egocentric spatial reasoning tasks, \textit{i.e.,} allocentric view orientation and scene simulation, which are already difficult in conventional 2D images~\citep{li2025viewspatial}. In the ODI setting, these tasks pose an even greater challenge, with model performance only marginally surpassing (or even falling below) that of the Blind GPT-4o baseline, suggesting that current models still fall short in capturing the spatial information conveyed by omnidirectional images.
\vspace{-9pt}
\subsection{Closed versus Open Evaluation}\label{sec:analysis}
\vspace{-6pt}
Comparing Table \ref{tab1-main} and Table \ref{tab3-open-ended}, we observe that model performance differs greatly between closed and open-ended QA settings. The findings demonstrate the necessity for conducting both closed and open-ended benchmarks. 
For tasks with a unique ground truth, \textit{i.e.,} counting and OCR, we can directly compare their performance across the two tables. The performance of these two tasks generally drops, indicating that {\textit{the choices may provide a hint for the MLLMs}}. Interestingly, {\textit{not all answer choices produce a positive effect}}. We observe cases where a model provides a correct response in the open-ended setting but selects the wrong option in the close-ended format. This discrepancy suggests that the presence of predefined options may sometimes introduce interference, and further reflects a potential difference between the model’s generative reasoning in open-ended tasks and its discriminative reasoning in multiple-choice tasks. 

We further observe that the model performance divergence is more evident in the open-ended setting, especially on spatial-level tasks, where models exhibit significantly larger variations. Unlike multiple-choice questions, where the given options constrain the answer space, open-ended responses can better reveal the differences between MLLMs’ reasoning and that of humans. For example, in the egocentric view orientation task, when no explicit constraints are imposed, models rarely produce {ego orientation terms}. Instead, they tend to describe orientations in relative terms. However, even when explicitly instructed to output absolute orientations, the models’ performance remains unsatisfactory. This suggests that MLLMs do not naturally reason about immersive ODI scenes in a human-like manner, \textit{i.e.,} by first engaging in perspective-taking and then conducting relative spatial analysis. Instead, their reasoning still resembles processing a warped 2D image.
\vspace{-9pt}
\section{Omni-CoT: Improving MLLMs Understanding of ODIs}
\vspace{-9pt}
In this section, we propose \textbf{Omni-CoT}, a training-free framework for improving MLLMs understanding capabilities on omnidirectional images by leveraging a human-like step-by-step chain-of-thought reasoning strategy, including viewpoint-guided answering, crop cue grounding and refining, and response refinement.
\vspace{-6pt}
\subsection{Framework Overview}\label{sec:omnicot}
\vspace{-3pt}
\subsubsection{Viewpoint Guided Answering}
\vspace{-3pt}
\input{figures/omni_cot}
Unlike conventional images, ODI comprehension requires MLLMs to extract viewpoint cognition from the projected ERP-format images. However, as analysed in Section~\ref{sec:analysis}, MLLMs often perceive ODIs merely as warped 2D images rather than reasoning within the immersive full-view setting, which poses a significant challenge even for proprietary models.
In 2D image comprehension, spatial understanding is often enhanced either through large-scale training or by incorporating prior information generated from 3D models~\citep{yang2025thinking, li2025viewspatial}. 
However, training-based approaches are resource-intensive and may overfit to the training data,
While 3D-derived features can provide useful cues, reliance on external models is often insufficient and not widely applicable.
To this end, We aim to explore training-free approaches to enhance MLLMs' understanding of ODIs by leveraging internal scene information through step-by-step reasoning.

A straightforward approach is to feed ODIs along with the multi-view images splitted from them into MLLMs, effectively guiding the models to view the ODIs in a human-like way by incorporating viewpoint information.
However, this approach is not practically feasible, as omnidirectional images inherently have high resolution, combining them with high-resolution multi-view inputs can easily exceed the model’s maximum input capacity, leading to failure. Moreover, high-resolution multi-view inputs generate a large number of image tokens, most of which are redundant or irrelevant, potentially hindering the model’s ability to focus on the critical information within the omnidirectional images.

To this end, we propose a more efficient approach by guiding MLLMs to explore the immersive environment presented by ODIs using compact textual prompts rather than additional image inputs, as presented in Figure \ref{figs:omni_cot}. Specifically, multi-view images are first extracted from the inverse sphere projection to generate six perspective views, \textit{i.e.,} top, bottom, right, left, front and back.
Subsequently, we use the MLLM to generate captions for each of the six viewpoints, capturing the key information contained in each view. By integrating these captions with the corresponding orientation information, the model can acquire a coarse understanding of the surrounding environment, thereby enhancing its global perception of the omnidirectional scene.
\subsubsection{Crop Cue Grounding and Refinement}
Directly extracting visual information from the full warped ERP image can be challenging for MLLMs. To address this, we propose a crop cue projection strategy. The MLLM is tasked with identifying the most relevant image crops, from which narrow-FoV crops are extracted as low-distortion visual cues to aid ODI comprehension. For a grounding box ($x_1,y_1,x_2,y_2$) where $(x_1, y_1)$ and $(x_2, y_2)$ represent the normalized coordinates of the top-left and bottom-right corners, respectively, the spherical parameters of the narrow-FoV crop, \textit{i.e.}, the spherical coordinates of the narrow-FoV cue's center $(\theta, \phi)$ and the approximate FoV $fov$ are computed as:
\begin{align}
\theta = -180^\circ + \frac{c_x}{W} \cdot 360^\circ&,  \phi = 90^\circ - \frac{c_y}{H} \cdot 180^\circ, \\
fov_w = (x_2 - x_1) \cdot 360^\circ&,  fov_h = (y_2 - y_1) \cdot 180^\circ, \\
fov = \mathrm{clip}\Big(\max(fov_w, &fov_h) + \mathrm{margin}, 30^\circ, 120^\circ\Big)
\end{align}
where ($c_x$, $c_y$) is the center of the crop, $W$ and $H$ are the width and height of the ERP image, respectively, \textit{margin} is an optional angular expansion to avoid overly tight cropping, and $\mathrm{clip}(\cdot)$ limits the FoV to a reasonable range.

However, since grounding may not always be accurate, relying solely on it can introduce unnecessary distractors. Therefore, we introduce a refinement mechanism, where the MLLM is queried again to label the relevance of each crop with respect to the question as ``yes'' or ``no'', and only the crops deemed relevant are fed back to the model to assist the response refinement step.
\subsubsection{Response Refinement}
\input{tabs/table4-experiment}
Finally, the ODI, along with the viewpoint captions, as well as the crop cues with their orientation information derived from the spherical coordinates, is fed back to assist in refining the response. The model is provided with its previous answer and prompted to rethink the answer based on the crop cues. As illustrated in Figure \ref{figs:omni_cot}.
\vspace{-6pt}
\subsection{Omni-CoT Performace}
\subsubsection{Performance Improvement on ODI-Bench}
\input{tabs/ablation_new}

\input{tabs/hyperparameter}
\input{tabs/table5-multiview}
\vspace{-6pt}
We conduct close-ended experiments on o3, GPT-4o, Gemini-2.0-flash, Qwen2.5-VL-72B and InternVL2.5-8B to validate the effectiveness of our framework on both proprietary and open-sourced MLLMs, the results are presented in Table \ref{tab4-exp}. As indicated in the table, viewpoint guiding serves as a key component for performance improvement, especially on spatial-related tasks. Moreover, the models performance could be overall further improved with the aid of crop cues.
\subsubsection{Ablation Studies}
We conduct ablation experiments to validate the effectiveness of each step in Omni-CoT, with results presented in Table \ref{tabs:ablation}. The results indicate that relying on direct grounding and cropping introduces unnecessary crops, which can degrade model performance. However, the proposed crop refinement step filters out irrelevant crops, thereby mitigating this negative effect and further boosting performance beyond the baseline with viewpoint guiding.
{Besides, viewpoint guiding serves as the fundamental component to improve model comprehension of spatial understanding, while crop grounding supplies the model with important visual cues for comprehensive understanding.
}

We further conduct hyperparameter ablations on GPT-4o to evaluate the effectiveness of the perspective FoV settings in Omni-CoT. As shown in Table~\ref{tabs:hyperparameter}, using a field of view of $90^{\circ}$ leads to the best overall performance, supporting the rationality of our chosen configuration.

\subsubsection{Comparison Experiments}
We provide additional baseline comparisons using both multi-view images and video-based inputs. For the multi-view setting, each omnidirectional image is projected into 12 perspective views (front, front-right, right, right-back, back, left-back, left, left-front, top-front, top-back, bottom-front, and bottom-back) and fed into the VLMs along with their orientation information. For the video setting, we convert each omnidirectional image into a 12-second, 60-FPS video that smoothly rotates through the front, right, back, left, and front views, followed by the top and bottom views. Besides, in Table 7 in the supplementary material, we compare the performance of Zero-shot CoT with our proposed Omni-CoT.

The results are presented in Table \ref{tab5-multiview}. Though multi-view and video-based inputs offer improvements on relative-direction and ODI reasoning tasks, they do not yield clear overall performance gains. Besides, their effectiveness is further limited by the models’ inherent constraints in handling multi-image or video inputs. While Zero-shot CoT provides minimal improvements on model performance, no clear improvement is observed in spatial-level tasks. In contrast, Omni-CoT consistently achieves the best results across all evaluated models, demonstrating its effectiveness and superiority over other inference pipelines. The experiment further highlights the necessity of our dedicated reasoning strategies tailored for omnidirectional image understanding.
\vspace{-6pt}
\section{Conclusion}
\vspace{-6pt}
In this work, we introduce ODI-Bench, a comprehensive benchmark for evaluating MLLMs’ ability to understand immersive environments presented by omnidirectional images. ODI-Bench consists of 2,000 high-quality omnidirectional images and over 4,000 question-answering pairs spanning 10 fine-grained tasks, covering both general-level and spatial-level understanding. We benchmark 20 leading MLLMs using both close-ended and open-ended evaluation settings. Our in-depth analysis of the experimental results reveals that current MLLMs still underperform on omnidirectional image understanding. To address this, we further propose Omni-CoT, a training-free approach to enhance MLLMs’ comprehension on omnidirectional images through chain-of-thought reasoning. Overall, ODI-Bench provides a rigorous yardstick for both evaluating and improving MLLMs on immersive environment understanding presented by ODIs.
\section*{Acknowledgement}
\vspace{-6pt}
This work was supported in part by the National Natural Science Foundation of China under Grants 62401365, 62225112, 62271312, 62132006, U24A20220, and in part by the China Postdoctoral Science Foundation under Grants BX20250411, 2025M773473.
\section*{Ethics Statement}
\vspace{-6pt}
Our work adheres to the ICLR Code of Ethics. This work introduces ODI-Bench, a benchmark for evaluating MLLMs understanding on omnidirectional images, and Omni-CoT, a training-free framework to enhance comprehension in immersive environments. All images in ODI-Bench are sourced from publicly available web resources under permissible use, with no personally identifiable information or sensitive data are included. Annotation was conducted by domain experts using VR devices, and only consensually verified QA pairs were retained to ensure annotation reliability. Our benchmark is intended purely for academic research and evaluation, and does not involve animal subjects, medical data, or applications in high-risk domains. We carefully considered potential societal impacts, including risks of misuse, and aim to provide the community with a transparent and rigorous tool for assessing VLMs’ capabilities in immersive environments, while promoting responsible stewardship of AI research.
\section*{Reproducibility statement}
\vspace{-6pt}
We have made efforts to ensure the reproducibility of our work.The main paper provides a detailed description of the construction process of ODI-Bench (Section \ref{sec:bench_process}), including task definitions and evaluation settings (Section \ref{sec:bench_exp}), and clearly outlines the design of the proposed Omni-CoT framework (Section~\ref{sec:omnicot}). In Appendix E, we provide the core prompts used in our experiments to facilitate replication. The dataset construction pipeline, annotation protocol, and evaluation methodology are all described in detail in the main text and supplementary materials. Together, these resources aim to provide sufficient transparency and guidance to reproduce our experimental results. We also plan to release the dataset and code upon publication.

\bibliography{iclr2026_conference}
\bibliographystyle{iclr2026_conference}

\newpage
\appendix
\section{Omnidirectional Image Viewing}
\input{figures/odi}
Captured by 360° cameras, ODIs provide a full field of view, which is significantly wider than that of a pinhole camera. Consequently, ODIs capture the entire surrounding environment with richer spatial information compared to traditional planar images. Due to their capability of delivering immersive experiences and complete perspectives, ODIs have been widely applied in various domains such as augmented reality (AR), virtual reality (VR), autonomous driving, and robotic navigation. In practice, raw ODI data are typically represented using equirectangular projection (ERP) or cubemap projection (CP) to ensure consistency with imaging pipelines. As a new data modality, ODIs possess both unique advantages (wide field of view enabled by spherical imaging, rich geometric cues, and flexible projection formats) and challenges (severe distortions in ERP and content discontinuity in CP). These characteristics make panoramic vision research both valuable and challenging. In this paper, we mainly focus on ERP-based ODIs, which are the most common format in real-world applications.

ERP provides a direct and simple way of representation. However, due to non-uniform sampling, scene objects are distorted depending on their relative positions in the image, with severe stretching near the poles of the sphere. In real-world applications, ERP-formated omnidirectional images are typically viewed through VR headmounted devices to create a fully immersive browsing experience, as shown in Figure~\ref{figs:odi}. Based on spherical coordinate projection with the $FoV=90^{\circ}$, one can roughly localize the viewer’s front, back, left, right, top, and bottom orientations. This viewing mode is significantly different from that of 2D viewing, thereby introducing challenges for ODI understanding.

\section{More Details of ODI-Bench}
\input{figures/all_tasks}
\input{figures/all_tasks_spatial}
In this section, we provide more details of our ODI-Bench. We provide one example of each tasks to better illustrate task in Figure \ref{figs:all_tasks} and Figure \ref{figs:all_tasks_spatial}.
\subsection{Detailed Task Descriotion}
\paragraph{Object Attribute}
This task challenges the MLLMs to localize the asked instance from high-resolution warped images and extract the attribute features from it. This task mainly focus on object-level attribute, \textit{e.g., }color, shape, material, functions, \textit{etc.}
\paragraph{Human Attribute}
This task challenges MLLMs beyond surface-level attribute recognition by requiring an understanding of human emotions. It encompasses human action recognition, human pose estimation, and even more demanding aspects such as human emotion recognition, \textit{etc}.
\paragraph{Existence}
This task evaluates a MLLM’s ability to recognize the existence of objects within high-resolution, complex spatial environments with massive instances. Unlike attribute-related questions, it requires the model to reason about the absence of objects. In our experiments, we observe that MLLMs generally perform poorly on existence-related questions.
\paragraph{Counting}
Unlike conventional 2D images, ODI counting poses greater challenges as it requires MLLMs to count in a full immersive view, which introduces two major challenges. First, identifying instances from warped viewpoints is inherently difficult, especially with splitted back views introduce distracting information. Second, the omnidirectional environment contains an overwhelming amount of visual information, making the counting task substantially more challenging than in standard 2D images.
\paragraph{OCR}
Given the inherent characteristics of ODI images, we aim to examine whether the OCR task presents distinct or greater challenges for MLLMs. We design this task as a general evaluation of MLLMs’ capability in ODI understanding. The primary difficulties arise from the need to comprehend high-resolution imagery, accurately ground the referenced text, and to handle viewpoint-induced hallucinations, as illustrated in the example shown in Figure \ref{figs:all_tasks}.
\paragraph{Egocentric View Orientation}
Conventional 2D images or NFoV videos only expose viewers to the frontal perspective. In contrast, browsing omnidirectional images inherently requires a 360-degree viewing experience, demanding that MLLMs browse the scene in a human-like manner. The most direct manifestation of this requirement is to specify the orientation of an object (front, back, left, right, up, down) from the viewer’s perspective. This constitutes the most fundamental distinction between omnidirectional image and 2D visual understanding. To this end, we propose the egocentric view-orientation task to assess whether MLLMs are capable of interpreting omnidirectional images within an immersive spatial context.
\paragraph{Allocentric View Orientation}
Spatial understanding from the allocentric (\textit{i.e.,} the viewpoint of other agents or a third-person depicted in the image) has been proven challenging in 2D images \citep{li2025viewspatial}. In this paper, we seek to evaluate whether the task poses greater challenge in ODI-presented immersive environment. Thus, we propose the Allocentric View-orientation task, in which the MLLM is required to determine the spatial orientation of instances relative to another person in the scene, effectively reasoning from the perspective of that individual rather than the viewer’s own point of view.
\paragraph{Scene Simulation}
Unlike egocentric or a third-party viewpoints, understanding from a virtual perspective requires higher spatial imagination ability. In this task, we specify the orientation and facing direction of a designated agent and ask the model to determine the spatial locations of other instances in the scene. Successfully completing this task requires the model to integrate abilities of spatial localization, directional reasoning, and immersive spatial understanding.
\paragraph{Relative Direction}
This task requires models to reason about relative spatial relations from an egocentric perspective, which not only demands accurate recognition and localization of objects, but also necessitates spatial reasoning within an immersive omnidirectional environment. Compared to conventional images, the projection distortions inherent to ERP format and the occlusion of back views make this task substantially more challenging.
\paragraph{ODI Reasoning}
This task challenges MLLMs beyond pure recognition or perspective taking tasks. Given a omnidirectional image, the model must engage in omnidirectional reasoning within a immersive environment to extract motion trajectories, infer spatial relationships, and even anticipate future actions. Such requirements place substantial demands on the model’s ability to integrate perception with higher-level reasoning about ODI environments.
\section{More Details of Benchmark Construction}
\input{figures/wordcloud}
Using the pipeline proposed in the main paper, we generated over 10,000 instance-level QA pairs, including human attribute and object attribute. Although the pipeline produces high-quality QA pairs, its generation strategy is based on single-viewpoint descriptions and only considers the segmented instance alone. Given the richness of omnidirectional imagery, such descriptions may fail to refer to a unique instance, even with viewpoint constraints, which undermines the reliability of the QA pairs. In addition, not all questions are semantically meaningful, some of them can be answered using common sense knowledge and therefore need to be filtered out. To address these issues, we employed manual filtering to ensure that the retained QA pairs are both uniquely referential and meaningful, also maintaining diversity within the instance-level dimension. After filtering, we preserved 1,220 object-attribute questions spanning over color, shape, material, texture, arrangement, \textit{etc.,} and 304 human-attribute questions including clothing, emotion, action, \textit{etc.}

Three domain experts were invited to annotate the QA pairs for ODI-bench except for the instance-level tasks. The annotation is conducted using VR headmounted displays rather than direct 2D viewing to ensure reliability and applicability. The annotation process lasted for one month. A cross-check was conducted afterwards, where only the QA pairs voted as correct by all experts were retained.

In the choice generation process, the omnidirectional image along with the QA pairs are fed into VLM to generate plausible choices, rather than inputting the answer alone. Manual examination is then performed to ensure choice quality. Specifically, the options should be reasonable, with no semantically incorrect ones. Besides, the choices should align with the correct answer to yield challenging distractors. Moreover, only one unambiguous correct option is permitted. Unqualified options were manually corrected. For direction-related options, three distractors were selected at orientations orthogonal to the ground-truth direction (\textit{i.e.}, $\pm90^{\circ}$ and $180^{\circ}$), so as to avoid ambiguous options. 

\section{Additional Experiments}
\input{figures/error_analysis}
\input{figures/zero_shot_omni_cot}
\subsection{Error Case Analysis}
To gain deeper insights into the bottlenecks of MLLMs for omnidirectional image understanding, we conduct detailed error case analyses based on the close-ended and open-ended evaluation results.
\subsubsection{Insufficient immersive cognition}
As the benchmark results indicate, MLLMs struggle to comprehend the immersive environments captured by omnidirectional images. Specifically, viewpoint information contributes little to their understanding of ODIs, resulting in a noticeable performance decline. As illustrated in Figure \ref{figs:error_case}, the models often respond using relative directions or interpret the ODI as a conventional 2D image, rather than employing orientation terms, even when explicitly instructed to do so. The finding indicates that MLLMs fail to ``view'' the ODIs like humans do, which poses challenges for ODI understanding. Importantly, this limitation not only hinders spatial-level reasoning but also affects other tasks requiring holistic immersion, such as counting and OCR, as presented in Figure \ref{figs:zero_shot_omni_cot}.
\subsubsection{Options as a Hint}
Choices serve as a strong prompt for models. Compared with open-ended responses, close-ended evaluation constrains the answer space, which inherently reduces uncertainty. Even when the options are designed to be distracting, they may not target the model’s weak spots, allowing the model to guess the correct answer without truly understanding the content. This short cuts is revealed under open-ended evaluation setting, as presented in Figure \ref{figs:error_case} (b).
\subsubsection{Option Anchoring Bias}
It is common for models to answer correctly in the close-ended setting but fail in the open-ended setting, as the latter presents a much larger solution space. However, we identify an intriguing phenomenon: while the model can provide correct answers in the open-ended setting, it sometimes fails in the close-ended setting by selecting the wrong option, as illustrated in Figure~\ref{figs:error_case} (c), we term this phenomenon as ``Option Anchoring Bias''. This observation highlights the necessity of evaluating both close-ended and open-ended settings. Specifically, the close-ended setting examines the model’s ability to ground its reasoning into predefined linguistic expressions, whereas the open-ended setting evaluates the model’s capability to respond in a manner consistent with human immersive viewing of ODIs. These two settings thus complement each other, and a comprehensive assessment should consider both perspectives for a holistic evaluation of MLLMs.

\subsection{Can Thinking Lead to Better Performace?}
\input{tabs/zero-shot-cot}
\input{tabs/openended-omni-cot}
Following Zero-shot-CoT~\citep{kojima2022large, wei2022chain}, we investigate how eliciting step-by-step reasoning from the MLLMs influences model performance on the omnidirectional image understanding tasks. The prompt is presented in Figure \ref{figs:zero-shot-cot}, we adopt the response format from \citep{guo2025deepseek} for better thinking and answering extraction. Experiments are conducted with GPT-4o and Gemini-2.0-flash, and the results for direct answering, Zero-shot CoT, and our Omni-CoT are summarized in Table~\ref{tabs:zero-shot-cot}. Qualitative results are presented in Figure \ref{figs:zero_shot_omni_cot}.

Compared to direct answering, Zero-shot CoT does slightly improve model performance, especially in the existence task. However, the improvement is minimal, and no clear improvement is observed in the spatial-level tasks, which further demonstrate that the model could not effectively utilize the immersive environment information. In comparison, our proposed Omni-CoT effectively boost model performance, especially in the spatial-level tasks, highlighting the effectiveness of our proposed framework.

\subsection{Filtering Ratio of Crop Refinement Step}
\input{tabs/Filter_ratio}
Filtering ratio per task of crop refinement step, a core step in Omni-CoT on GPT-4o is reported in Table \ref{tabs:Filter_ratio}. Across all tasks, this step successfully removes a substantial amount of redundant information. Together with the results in Table~\ref{tabs:ablation}, these findings demonstrate that crop refinement significantly contributes to the overall performance improvement.

\subsection{Open-ended Evaluation Results of Omni-CoT}
We present Omni-CoT on open-ended evaluation setting in Table \ref{tabs:omni-cot-openeded} to further demonstrate the effectiveness of the proposed framework.  InternVL2.5-8B and Gemini-2.0-flash are adopted for experiment. Even under unconstrained open-ended setting, Omni-CoT significantly enhances model performance on an overall undeerstanding of omnidirectional images. In particular, Gemini-2.0-flash shows a remarkable improvement of 33.14\% on egocentric view orientation task. These results further demonstrate that Omni-CoT greatly improves comprehension on immersive environments presented by ODIs.

\input{tabs/resize}
\input{tabs/multiple_runs}
\input{tabs/prompts}
\subsection{Discussion on High-resolution Images}
Some recent research~\citep{yang2025visionthink, liao2025we} have discussed that in many general scenarios, simple resizing can achieve strong performance. This raises the question of whether such a straightforward technique would also be effective on our ODI-Bench, given that most images in our benchmark are of high resolution. To investigate this, we downsample all images to a resolution of $512\times 1024$ (comparable to that used in previous ODI benchmarks) before feeding them into the MLLMs. We evaluate two strong baseline models, InternVL3-78B and GPT-4o. The corresponding results are presented in Table~\ref{tabs:resize}.

For general-level tasks, model performance drops substantially after resizing. This is expected because in the information-dense 360 view, the instances relevant to general tasks typically occupy only a small portion of the entire image. Consequently, resizing blurries these fine-grained regions, leading to significant information loss. This phenomenon is consistent with prior findings in 2D image settings, where OCR-related tasks were shown to be most affected by reductions in input resolution. 

In contrast, spatial-level tasks are less affected by token reduction. We hypothesize that these tasks rely more on capturing the global spatial layout rather than fine-grained instance details, allowing the models to maintain relatively stable performance even under reduced image resolution. 
\subsection{Performance over Multiple Runs}
We evaluate GPT-4o and InternVL2.5-8B on ODI-Bench by running each model three times and reporting the mean and standard deviation of their performance. The results are presented in Table \ref{tabs:multiple_runs}. The performance variation remains within a minimal and acceptable range, further demonstrating the reliability and stability of our benchmark.

\subsection{Influence of Prompt}
Considering the crucial role of prompts in guiding model understanding, we further explore how different prefix instructions influence model performance. To this end, we experiment with several prompt variants to assess their effects on model performance on ODI-Bench. The results are summarized in Table~\ref{tabs:prompt}.
The performance variation across different prompts remains within a small and acceptable range. Moreover, even when provided with more explicit and descriptive instructions, the models show no substantial improvement on spatial-level tasks. This indicates that relying solely on prompt engineering is insufficient for enabling models to genuinely understand omnidirectional content.
\subsection{Inference Time}
\input{tabs/inf_time}
Inference time comparison for direct answering, Zero-shot CoT, Omni-CoT (with only viewpoint guiding), and the full Omni-CoT pipeline is reported in Table \ref{tabs:inf_time}. The experiments are conducted on both InternVL2.5-8B and o3. As shown in the results, Omni-CoT (only w/ viewpoint guiding) incurs only a slightly higher inference time than Zero-shot CoT, yet delivers substantially better overall performance. Furthermore, the full Omni-CoT pipeline brings additional performance gains with a reasonable increase in computation cost.
\section{Prompts}\label{prompts}
\input{figures/prompt_cot}
\input{figures/prompt_exp}
In this section, we present the prompts used throughout the benchmark construction and benchmark experiment.
\subsection{Benchmark Construction Prompts}
\input{figures/prompt_qa}
Prompts are adopted in the automatic data curation pipeline as well as the distractor generation process. The prompts are as presented in Figure \ref{figs:prompt_qa}.
\subsection{Benchmark Prompts}
Prompts adopted in the close-ended and open-ended benchmark experiment are shown in Figure \ref{figs:prompt_exp}. For close-ended evaluation, slightly different prompts are employed to ensure the output format of the responses. For open-ended evaluation, due to the fact that models may overlook the immersive environmental perspective and instead tend to answer in a relative direction manner, we explicitly require them to answer with an [orientation word], forcing them to respond starting with a egocentric view orientation expression.
\subsection{LLM Evaluator Prompts}\label{sec:llm_evaluator}
\input{figures/prompt_llm}
For open-ended responses, we adopted LLM-based evaluator to score the response between 0 to 1. Specifically, we leverage GPT-4o to assist evaluation using a instruction-based prompting approach. As detailed in Figure \ref{figs:prompt_llm}, we cover examples that are fully correct (\textit{i.e.}, 1.0) or incorrect (\textit{i.e.}, 0.0), as well as examples used to define different types of “partially correct” responses.
Different evaluation prompts are adopted based on tasks.

For answers with a unique ground truth, \textit{i.e.,} counting and existence, only the key answers strictly aligns with the output is rated as ``correct'', else rated as ``incorrect''. However, for OCR task, we do not adopt this binary scheme. Instead, answers with correctly recognized characters but in an incorrect order are assigned a partial score of 0.5, in order to further distinguish the model’s OCR capability.
While for attribute-level responses, the answers are rated based on the similarity between the key content and the ground truth. In addition, the original question is also taken into account when judging the model’s responses, so as to ensure a more accurate and context-aware evaluation.

For tasks related to direction-based output, \textit{e.g.,} egocentric view orientation and relative direction, \textit{etc.}, we adopt a instruction-based prompting strategy to guide the scoring process. Specifically, if the model’s prediction deviates from the ground-truth orientation by 45 degrees, a partial score of 0.5 is assigned; if it is completely correct, a full score of 1 is given; otherwise, the score is 0.
\subsection{Omni-CoT Prompts}
\input{figures/prompt_omnicot}
Prompts for our proposed Omni-CoT are presented in Figure \ref{figs:prompt_omnicot}.

\section{Broader Impacts}
ODI-Bench provides a promising direction for real-world applications by comprehensively evaluating MLLMs on omnidirectional image understanding. By assessing both general-level and spatial-level understanding, we aim to contribute to the advancement of MLLMs towards more immersive and context-aware perception. The general-level tasks, such as attribute recognition, existence, and OCR, are closely aligned with key applications of omnidirectional vision in autonomous driving, VR/AR-based human–computer interaction, and robotic navigation. The spatial-level tasks encourage accurate orientation reasoning, supporting safety-critical applications like navigation and autonomous driving. In particular, the allocentric view orientation and scene simulation tasks contribute to embodied intelligence research, as they equip agents with the ability to reason about unseen viewpoints, reconstruct spatial layouts, and simulate the environment for planning and decision-making.

Beyond serving as an evaluation benchmark, the the richness of our data and the diversity of question types also make ODI-Bench a valuable resource for pre-training and post-training of MLLMs. Leveraging ODI-Bench can effectively enhance models’ understanding of immersive environments and improve their generalization capabilities in such settings.
\section{Data License}
All images in our benchmark are sourced from Flickr under licenses that permit redistribution and academic use. In accordance with these licenses, we only include images that are legally allowed for reuse in research settings. The entire benchmark will be released under the CC BY 4.0 license, enabling free use and redistribution with proper attribution.
\section{LLM Usage Statement}
Large Language Models (LLMs) were only used as an assistive tool for minor language polishing and improving readability of the manuscript. All aspects of research ideation, benchmark construction, method design, implementation, data collection, and analysis were conducted entirely by the authors. No part of the research design, experimental process, or analysis was generated by LLMs. The authors take full responsibility for all content of this paper.
\end{document}

%% file: math_commands.tex
\usepackage{amsmath,amsfonts,bm}

\def\eqref#1{equation~\ref{#1}}

\def\1{\bm{1}}

\DeclareMathAlphabet{\mathsfit}{\encodingdefault}{\sfdefault}{m}{sl}
\SetMathAlphabet{\mathsfit}{bold}{\encodingdefault}{\sfdefault}{bx}{n}



%% file: figures/teaser.tex
\begin{figure}[h]
\vspace{-2.5em}
\begin{center}
\includegraphics[width=0.95\linewidth]{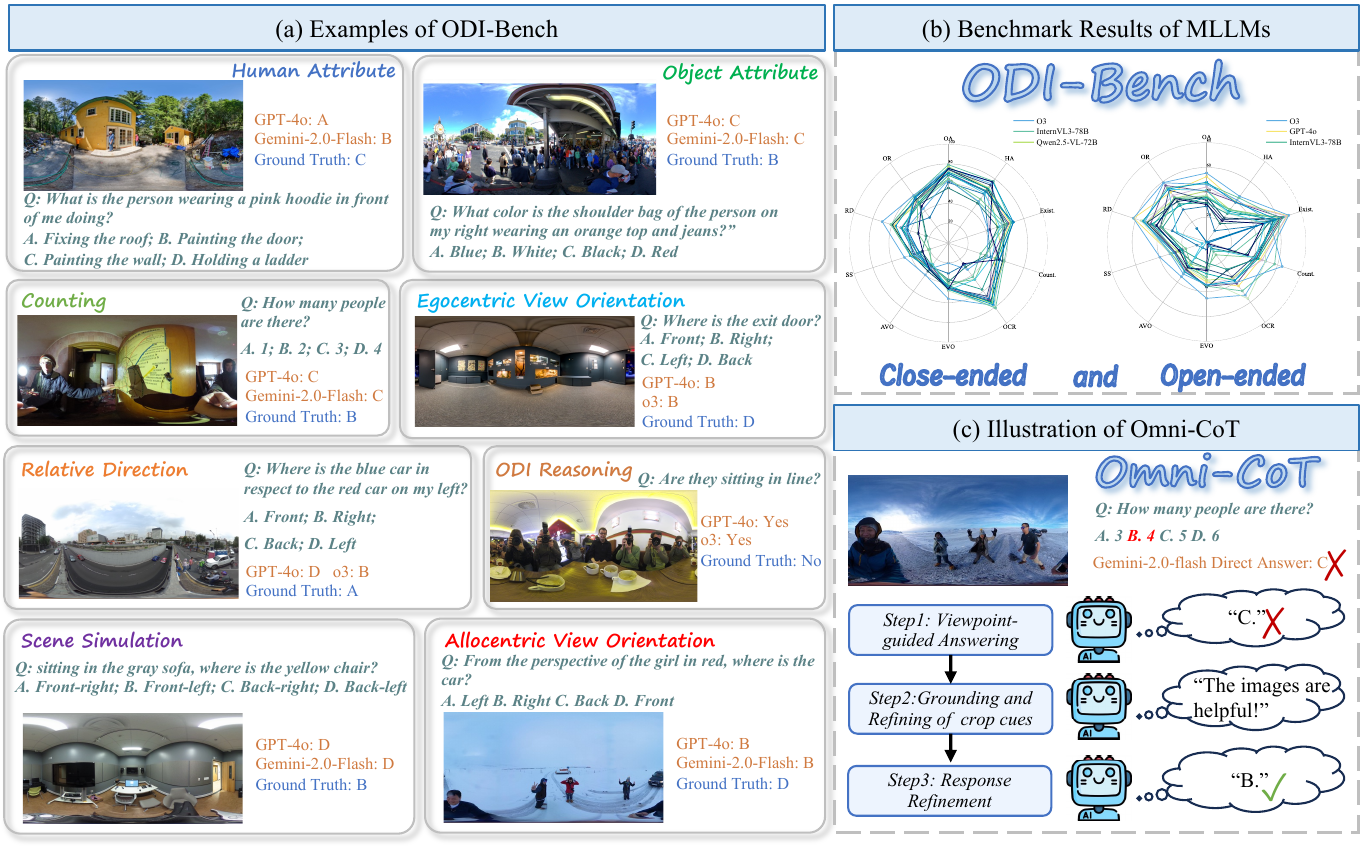}
\end{center}
\vspace{-1.5em}
\caption{
{(a) We introduce \textbf{ODI-Bench}, a comprehensive benchmark for omnidirectional image understanding, covering \textit{10} diverse tasks. (b) 20 leading MLLMs are benchmarked with both close-ended and open-ended evaluation. (c) To further improve model performance, we propose \textbf{Omni-CoT}, a chain-of-thought framework that enhances MLLMs’ comprehension on omnidirectional images via step-by-step reasoning.}}
\label{figs:teaser}
\end{figure}

%% file: tabs/table2-comparison.tex
\begin{table}
\vspace{-2em}
\setlength{\belowcaptionskip}{-0.01cm}
\centering
\belowrulesep=0pt
\aboverulesep=0pt
\renewcommand\arraystretch{1.2}
\caption{Comparison between widely adopted general benchmarks, omnidirectional benchmarks, and our ODI-Bench. The first row group presents commonly used image benchmarks, the middle row group includes two spatial benchmarks, and the last row group lists ODI benchmarks.}
   \resizebox{\linewidth}{!}{
   \begin{tabular}{l|ccc|cccccccc}
    \toprule[1pt]
     \multirow{2}{*}{Benchmark} & \multirow{2}{*}{\#Images} & \multirow{2}{*}{\#QA Pairs} & \multirow{2}{*}{\#Question Type} & \multirow{2}{*}{Visual Modality} & \multirow{2}{*}{Max Reso.}& \multirow{2}{*}{Real Scenes}  & \multicolumn{2}{c}{Evaluation} & \multicolumn{2}{c}{Dimension} & \multirow{2}{*}{QA Source} \\
\cmidrule(lr){8-9} \cmidrule(lr){10-11}
 & & & & & & & Open & Close & General & Spatial & \\
     \midrule
     MMBench~\citep{liu2024mmbench}&3,217&3,217&20&2D image&$<$1K&\textcolor{green}{\ding{51}}&\textcolor{red}{\ding{55}}&\textcolor{green}{\ding{51}}&\textcolor{green}{\ding{51}}&\textcolor{red}{\ding{55}}&Manual\\
     MM-Vet~\citep{yu2023mm}&200&218&16&2D image&$<$6K&\textcolor{green}{\ding{51}}&\textcolor{green}{\ding{51}}&\textcolor{red}{\ding{55}}&\textcolor{green}{\ding{51}}&\textcolor{green}{\ding{51}}&Manual\&Existed\\
     \midrule
     ViewSpatial-Bench~\citep{li2025viewspatial}&1,000&5,700&5&3D Scene&$<$1K&\textcolor{green}{\ding{51}}&\textcolor{green}{\ding{51}}&\textcolor{red}{\ding{55}}&\textcolor{red}{\ding{55}}&\textcolor{green}{\ding{51}}&Auto\\
     VSI-Bench~\citep{yang2025thinking}&29&5,000&8&NFOV Video&$<$1K&\textcolor{green}{\ding{51}}&\textcolor{green}{\ding{51}}&\textcolor{green}{\ding{51}}&\textcolor{red}{\ding{55}}&\textcolor{green}{\ding{51}}&Auto\\
     SSRBench~\citep{liu2025ssr}&789&789&6&2D Image&-&\textcolor{green}{\ding{51}}&\textcolor{green}{\ding{51}}&\textcolor{green}{\ding{51}}&\textcolor{green}{\ding{51}}&\textcolor{green}{\ding{51}}&Auto\\
     \midrule
     VQA 360$^{\circ}$~\citep{chou2020visual}&1,490&17,000&6&Indoor ODI&1K&\textcolor{red}{\ding{55}}&\textcolor{red}{\ding{55}}&\textcolor{green}{\ding{51}}&\textcolor{green}{\ding{51}}&\textcolor{green}{\ding{51}}&Auto\\
     OSR-Bench~\citep{dongfang2025multimodal}&4,100&153,000&3&Indoor ODI&1K&\textcolor{red}{\ding{55}}&\textcolor{green}{\ding{51}}&\textcolor{red}{\ding{55}}&\textcolor{red}{\ding{55}}&\textcolor{green}{\ding{51}}&Auto\\
     Dense360-Bench~\citep{zhou2025dense360}&1,279&6,000&2&Indoor\&Outdoor ODI&-&\textcolor{green}{\ding{51}}&\textcolor{green}{\ding{51}}&\textcolor{red}{\ding{55}}&\textcolor{green}{\ding{51}}&\textcolor{red}{\ding{55}}&Auto\\
     {ODI-Bench} (Ours)&2,000&4,254&10&Indoor\&Outdoor ODI&12K&\textcolor{green}{\ding{51}}&\textcolor{green}{\ding{51}}&\textcolor{green}{\ding{51}}&\textcolor{green}{\ding{51}}&\textcolor{green}{\ding{51}}&Manual\&Auto\\
    \bottomrule[1pt]
  \end{tabular}}
  \label{tab2:comparison}
  \vspace{-7mm}
\end{table}

%% file: figures/distribution.tex
\begin{figure}[t]
\vspace{-1em}
\begin{center}
\includegraphics[width=\linewidth]{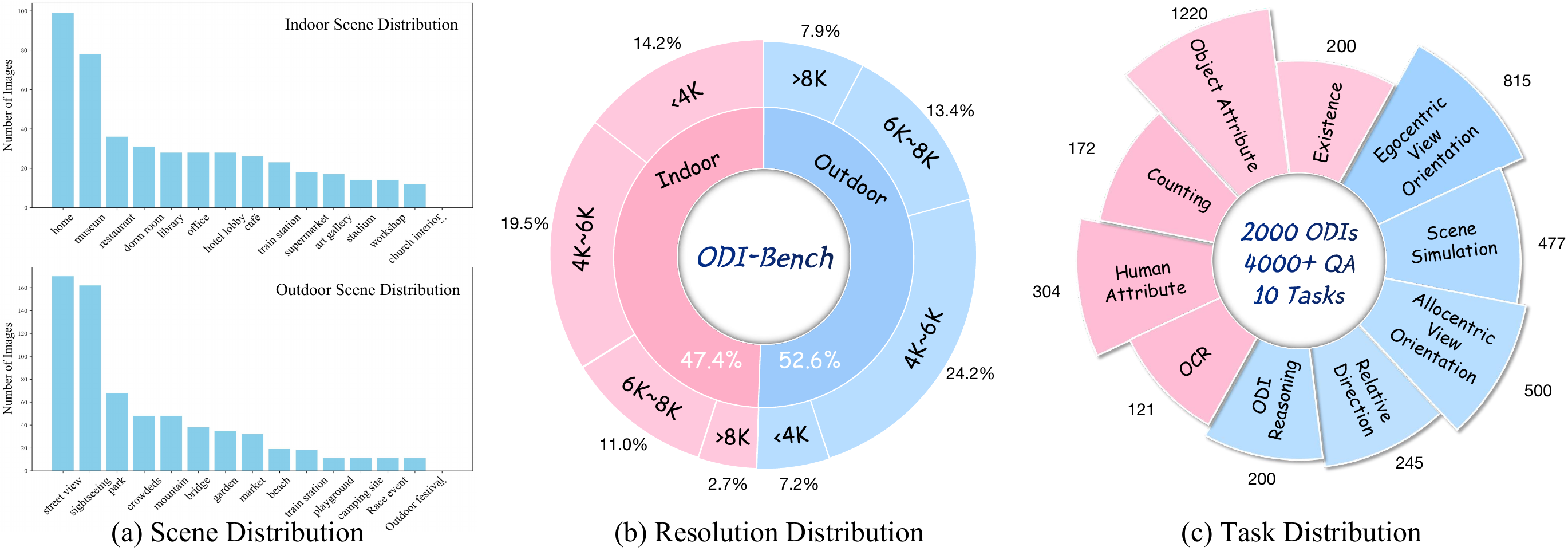}
\end{center}
\vspace{-1.2em}
\caption{Data distribution in ODI-Bench.}
\label{figs:distribution}
\vspace{-5mm}
\end{figure}

%% file: figures/pipeline.tex
\begin{figure}[t]
\vspace{-1.5em}
\begin{center}
\includegraphics[width=\linewidth]{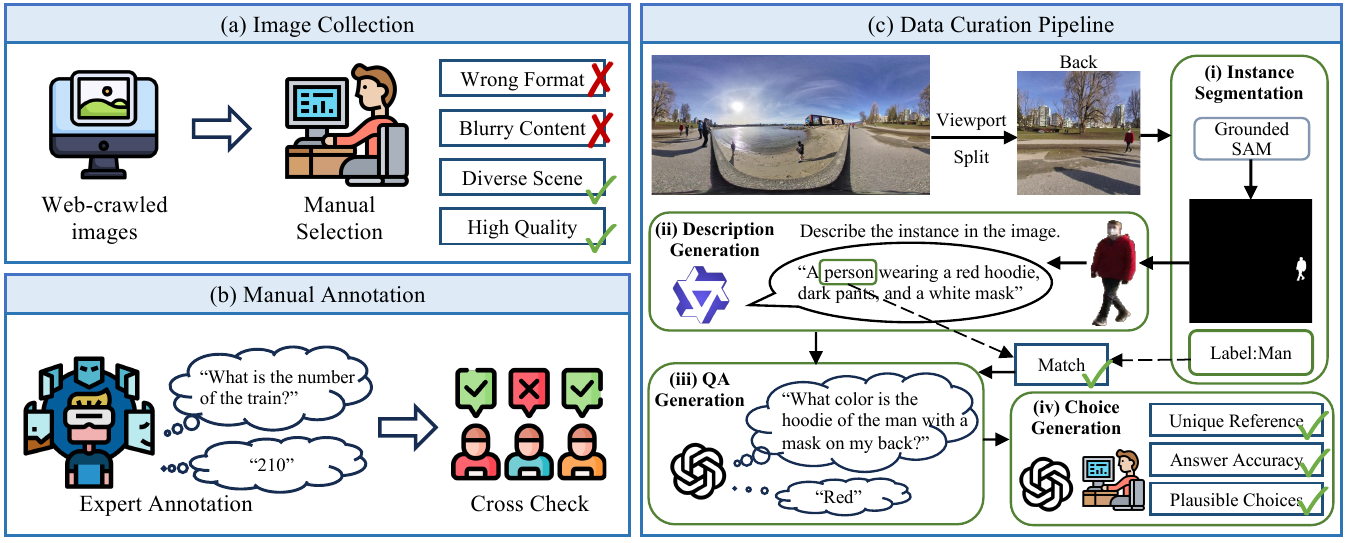}
\end{center}
\vspace{-1.5em}
\caption{Construction procedures of ODI-Bench. (a) The benchmark images are carefully selected to ensure quality and diversity. (b) The majority tasks are manually annotated by human experts. (c)
{Object Attribute and Human Attribute QA pairs are generated through a dedicated annotation pipeline with human verification to guarantee quality.} }
\label{figs:pipeline}
\vspace{-7mm}
\end{figure}

%% file: tabs/table1-main.tex
\begin{table}[t]
\vspace{-2em}
\setlength{\belowcaptionskip}{-0.01cm}
\centering
\belowrulesep=0pt
\aboverulesep=0pt
\renewcommand\arraystretch{1.2}
\caption{Benchmark results for MLLMs under the close-ended evaluation setting.The tasks are defined as follows: OA (Object Attribute), HA (Human Attribute), Exist. (Existence), Count. (Counting), EVO (Egocentric View Orientation), AVO (Allocentric View Orientation), SS (Scene Simulation), RD (Relative Direction), OR (ODI Reasoning). For each task, the \textbf{best-performance} is indicated in bold and the \underline{second-best} are underlined.}
\footnotesize
   \resizebox{\linewidth}{!}{
   \begin{tabular}{
p{5.8cm} 
*{11}{>{\centering\arraybackslash}p{0.90cm}} 
}
    \toprule[1pt]
     &\multirow{2}{*}{\textit{Overall}}&\multicolumn{5}{c}{\textit{General}}&\multicolumn{5}{c}{\textit{Spatial}}\\
  \cmidrule(lr){3-7} \cmidrule(lr){8-12}
  &&OA&HA&Exist.&Count.&OCR&EVO&AVO&SS&RD&OR\\
    \midrule
    \rowcolor{myblue!40}\multicolumn{12}{c}{\itshape Proprietary MLLMs}\\
    \midrule
    GPT-4o~\citep{hurst2024gpt}&55.79&74.43&67.76&65.50&49.42&74.38&43.24&32.49&39.60&57.55&53.50\\
    Qwen-VL-Plus~\citep{bai2025qwen2}&53.85&\underline{74.51}&68.75&43.00&\underline{52.33}&67.77&46.13&29.98&34.80&50.61&49.00\\
    Gemini-2.0-flash~\citep{comanici2025gemini}&\underline{57.12}&73.03&\underline{69.41}&\underline{66.50}&52.33&\textbf{80.16}&\underline{48.10}&\underline{32.91}&\underline{40.20}&\underline{61.13}&\underline{54.00}\\
    o3~\citep{openai2025o3systemcard}&\textbf{62.62}&\textbf{75.82}&\textbf{74.01}&\textbf{75.00}&\textbf{57.56}&\underline{77.69}&\textbf{56.20}&\textbf{39.62}&\textbf{46.60}&\textbf{70.20}&\textbf{59.50}\\
    
    \midrule
    \rowcolor{myblue!40}\multicolumn{12}{c}{\itshape Open-sourced MLLMs}\\
    \midrule
    
    LLaVA-v1.5-7B~\citep{liu2024improved}&45.04&64.02&51.44&58.00&34.30&42.15&43.86&28.93&19.60&24.08&50.00\\
    LLaVA-ov-0.5B~\citep{li2024llavaonevision}&44.25&61.21&61.26&42.00&44.77&27.27&30.61&29.77&35.40&50.20&32.00\\
    idefics3-8B~\citep{laurençon2024building}&49.89&68.45&65.32&62.50&50.00&57.85&36.71&28.79&30.20&49.39&49.50\\
    XComposer2~\citep{dong2024internlm}&51.84&75.57&\underline{76.32}&44.00&54.07&25.62&46.56&32.29&27.20&31.84&46.00\\
    Deepseek-VL-1.3B~\citep{lu2024deepseek}&42.02&55.66&50.00&52.50&39.53&29.75&39.02&28.18&28.00&23.27&49.00\\
    LLava-Next-7B~\citep{li2024llava}&45.91&64.84&54.93&60.50&39.77&45.45&40.42&27.04&21.20&32.24&53.50\\
    LLava-Next-34B~\citep{li2024llava}&52.24&70.57&62.17&54.50&45.93&45.45&\textbf{47.42}&26.42&38.00&45.71&57.50\\
    glm-4v-9B~\citep{glm2024chatglm}&53.20&70.41&64.14&\textbf{69.00}&54.65&69.42&44.29&\underline{32.91}&30.80&43.27&57.50\\
    MiniCPM-V 4.0~\citep{yao2024minicpm}&53.71&71.72&68.42&\underline{67.00}&51.74&74.38&42.58&32.29&33.00&49.39&51.00\\
    InternVL2.5-8B~\citep{chen2024expanding}&52.76&68.52&70.07&60.00&51.74&66.12&45.23&31.24&33.00&44.08&\underline{58.00}\\
    
    Qwen2.5-VL-3B~\citep{bai2025qwen2}&52.88&70.66&65.13&64.00&45.93&71.07&46.13&28.60&33.00&45.71&53.50\\
    Qwen2.5-VL-32B~\citep{bai2025qwen2}&56.70&74.69&67.76&60.00&57.56&72.73&45.97&\textbf{35.01}&38.20&59.59&54.50\\
    Qwen2.5-VL-72B~\citep{bai2025qwen2}&56.91&\underline{77.38}&68.75&52.00&\underline{58.14}&75.21&\underline{46.75}&32.08&\underline{38.40}&\underline{60.00}&50.00\\
    InternVL3-38B~\citep{zhu2025internvl3}&\underline{57.91}&75.57&\underline{76.32}&\textbf{69.00}&54.07&\underline{76.86}&46.56&32.29&\textbf{40.40}&55.92&56.50\\
    InternVL3-78B~\citep{zhu2025internvl3}&\textbf{59.43}&\textbf{79.18}&\textbf{77.30}&66.50&\textbf{59.30}&\textbf{80.99}&46.01&31.67&\textbf{40.40}&\textbf{60.82}&\textbf{58.50}\\
    Intern-s1~\citep{bai2025intern}&42.37&63.93&51.97&42.50&40.94&33.06&20.15&32.08&31.00&45.31&43.00\\
    \midrule
    \rowcolor{myblue!40}\multicolumn{12}{c}{\itshape Baseline}\\
    \midrule
    Blind GPT-4o&36.39&58.93&42.43&17.00&17.44&29.75&21.96&29.14&31.60&29.80&25.50\\
    Random Choice&26.93&25.00&25.00&50.00&25.00&25.00&25.00&25.00&25.00&25.00&41.00\\
    \bottomrule[1pt]
  \end{tabular}}
  \label{tab1-main}
  \vspace{-7mm}
\end{table}

%% file: tabs/table3-openended.tex
\begin{table}[t]
\setlength{\belowcaptionskip}{-0.01cm}
\centering
\belowrulesep=0pt
\aboverulesep=0pt
\renewcommand\arraystretch{1.2}
\caption{Benchmark results for MLLMs  under the open-ended evaluation setting. For each task, the \textbf{best-performance} is indicated in {bold} and the \underline{second-best} are underlined.}
\footnotesize
   \resizebox{\linewidth}{!}{
   \begin{tabular}{
p{5.8cm} 
*{11}{>{\centering\arraybackslash}p{0.90cm}} 
}
    \toprule[1pt]
     &\multirow{2}{*}{\textit{Overall}}&\multicolumn{5}{c}{\textit{General}}&\multicolumn{5}{c}{\textit{Spatial}}\\
  \cmidrule(lr){3-7} \cmidrule(lr){8-12}
  &&OA&HA&Exist.&Count.&OCR&EVO&AVO&SS&RD&OR\\
    \midrule
    \rowcolor{myblue!40}\multicolumn{12}{c}{\itshape Proprietary MLLMs}\\
    \midrule
    GPT-4o~\citep{hurst2024gpt}&\underline{42.91}&\underline{52.62}&\underline{39.74}&\underline{68.50}&45.35&43.80&32.27&27.25&30.30&\underline{61.84}&\underline{49.50}\\
    Qwen-VL-Plus~\citep{bai2025qwen2}&39.87&44.39&{39.35}&67.50&\underline{47.00}&36.36&\underline{34.39}&\underline{29.10}&27.65&51.09&46.20\\
    Gemini-2.0-flash~\citep{comanici2025gemini}&36.42&37.82&28.55&48.26&\textbf{50.00}&\textbf{56.50}&30.86&25.89&\underline{31.00}&55.51&42.17\\
    o3~\citep{openai2025o3systemcard}&\textbf{49.53}&\textbf{55.49}&\textbf{45.36}&\textbf{69.50}&\textbf{50.00}&\underline{52.89}&\textbf{45.40}&\textbf{34.59}&\textbf{39.60}&\textbf{62.04}&\textbf{59.10}\\
    \midrule
    \rowcolor{myblue!40}\multicolumn{12}{c}{\itshape Open-sourced MLLMs}\\
    \midrule
    
    LLaVA-v1.5-7B~\citep{liu2024improved}&32.29&35.02&32.07&56.50&24.42&9.504&28.28&25.47&26.60&45.10&43.50\\
    LLaVA-ov-0.5B~\citep{li2024llavaonevision}&17.90&28.75&11.12&42.00&19.19&3.719&6.196&5.765&8.300&29.18&32.20\\
    idefics3-8B~\citep{laurençon2024building}&27.93&31.50&28.16&\textbf{65.00}&43.60&27.27&16.75&15.41&20.60&37.35&37.90\\
    XComposer2~\citep{dong2024internlm}&28.36&30.12&20.92&48.00&34.88&1.652&23.13&25.79&23.00&42.65&43.15\\
    Deepseek-VL-1.3B~\citep{lu2024deepseek}&27.80&31.33&18.22&53.50&33.14&6.198&20.98&19.92&23.10&42.24&44.15\\
    LLava-Next-7B~\citep{li2024llava}&30.52&36.58&32.24&60.50&40.12&12.81&21.04&22.01&17.60&38.78&44.45\\
    LLava-Next-34B~\citep{li2024llava}&35.25&41.57&29.28&49.50&41.52&7.025&35.40&20.21&24.60&45.51&52.42\\
    glm-4v-9B~\citep{glm2024chatglm}&35.79&38.80&26.12&60.50&42.44&32.23&36.38&25.26&26.50&41.43&42.95\\
    MiniCPM-V 4.0~\citep{yao2024minicpm}&32.52&36.02&30.56&57.00&43.60&38.43&29.34&24.32&21.04&46.53&36.25\\
    InternVL2.5-8B~\citep{chen2024expanding}&30.86&33.52&22.34&62.00&39.53&34.71&21.96&25.68&20.40&38.98&51.55\\
    Qwen2.5-VL-3B~\citep{bai2025qwen2}&38.91&40.80&39.64&\underline{63.50}&41.86&\underline{41.32}&\underline{35.77}&28.09&26.80&47.35&\underline{56.20}\\
    Qwen2.5-VL-32B~\citep{bai2025qwen2}&37.67&39.56&26.34&62.00&42.44&30.91&\textbf{36.47}&\underline{30.21}&\underline{30.43}&50.41&44.25\\
    Qwen2.5-VL-72B~\citep{bai2025qwen2}&39.49&45.89&35.14&63.00&42.44&39.55&34.91&24.10&27.60&\textbf{54.39}&47.75\\
    InternVL3-38B~\citep{zhu2025internvl3}&40.96&\textbf{48.11}&39.57&59.00&40.94&38.43&34.66&{29.77}&27.60&\underline{54.08}&52.62\\
    InternVL3-78B~\citep{zhu2025internvl3}&\textbf{42.52}&\underline{47.05}&\textbf{43.91}&62.00&\underline{57.56}&\textbf{48.76}&35.58&{29.77}&{29.82}&{53.27}&53.95\\
    
    Intern-s1~\citep{bai2025intern}&\underline{42.12}&46.30&\underline{41.68}&53.50&\textbf{63.95}&40.50&34.72&\textbf{31.87}&\textbf{30.80}&52.04&\textbf{58.85}\\

    \bottomrule[1pt]
  \end{tabular}}
  \label{tab3-open-ended}
  \vspace{-5mm}
\end{table}

%% file: figures/omni_cot.tex
\begin{figure}[t]
\vspace{-1.5em}
\begin{center}
\includegraphics[width=1\linewidth]{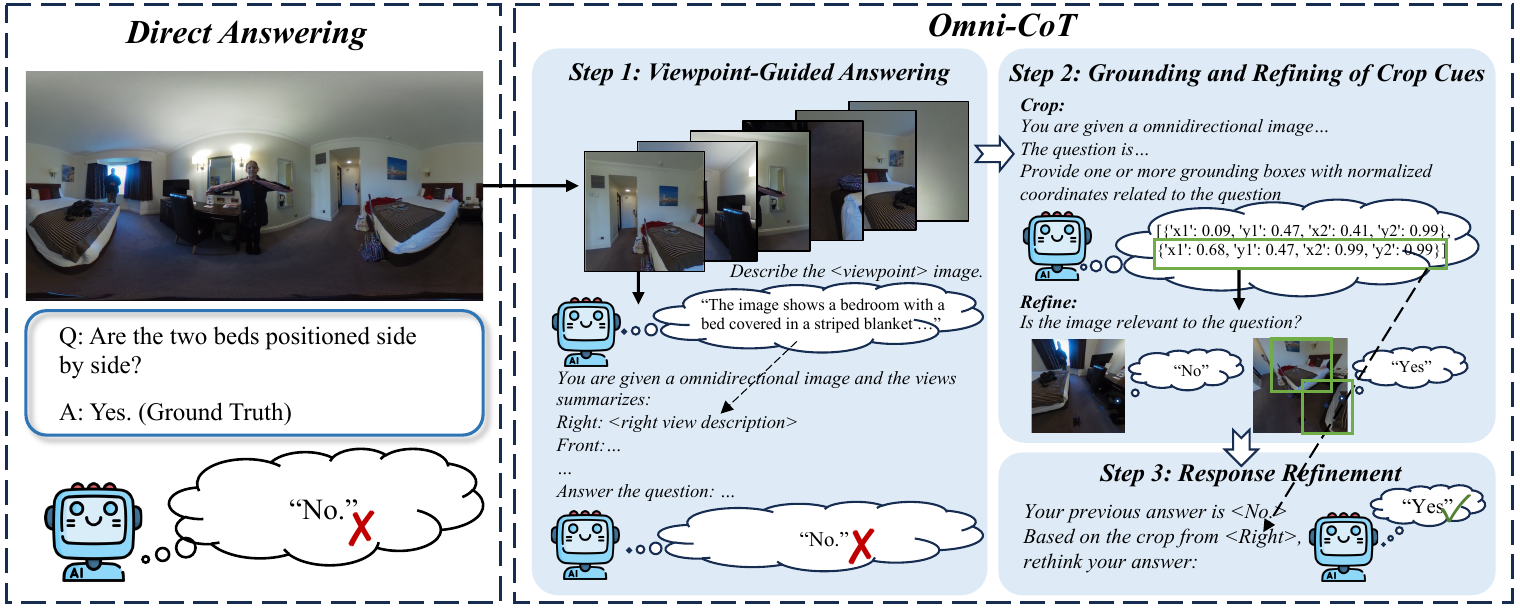}
\end{center}
\vspace{-1.5em}
\caption{We introduce \textbf{Omni-CoT}, The framework enhances VLMs’ comprehension of omnidirectional images via chain-of-thought reasoning through three steps: viewpoint-guided answering, grounding and refinement of crop cues, and response refinement. Compared with direct answering, Omni-CoT achieves notable performance improvements.}
\label{figs:omni_cot}
\vspace{-5mm}
\end{figure}

%% file: tabs/table4-experiment.tex
\begin{table}
\vspace{-2em}
\setlength{\belowcaptionskip}{-0.01cm}
\centering
\belowrulesep=0pt
\aboverulesep=0pt
\renewcommand\arraystretch{1.2}
\caption{Performance of \textbf{Omni-CoT} on ODI-Bench, better performances over baseline are \textbf{bolded}.}
\scriptsize
   \resizebox{\linewidth}{!}{
   \begin{tabular}{l
>{\centering\arraybackslash}p{0.85cm}
>{\centering\arraybackslash}p{0.85cm}
>{\centering\arraybackslash}p{0.85cm}
>{\centering\arraybackslash}p{0.85cm}
>{\centering\arraybackslash}p{0.85cm}
>{\centering\arraybackslash}p{0.85cm}
>{\centering\arraybackslash}p{0.85cm}
>{\centering\arraybackslash}p{0.85cm}
>{\centering\arraybackslash}p{0.85cm}
>{\centering\arraybackslash}p{0.85cm}
>{\centering\arraybackslash}p{0.85cm} 
}
    \toprule[1pt]
     \multirow{2}{*}{\textit{Method}}&\multirow{2}{*}{\textit{Overall}}&\multicolumn{5}{c}{\textit{General}}&\multicolumn{5}{c}{\textit{Spatial}}\\
  \cmidrule(lr){3-7} \cmidrule(lr){8-12}
  &&OA&HA&Exist.&Count.&OCR&EVO&AVO&SS&RD&OR\\
  \midrule
    \textbf{o3}&{62.62}&\textbf{75.82}&{74.01}&{75.00}&{57.56}&{77.69}&{56.20}&{39.62}&{46.60}&{70.20}&{59.50}\\
    \hdashline
    (w/ Viewpoint Guiding)&\textbf{68.78}&75.57&\textbf{75.66}&\textbf{78.00}&55.81&77.69&\textbf{81.23}&\textbf{41.93}&\textbf{54.80}&68.57&\textbf{62.00}\\
    $\Delta(\uparrow)$&+6.16&-0.25&+1.65&+3.00&-1.75&+0.00&+25.03&+2.31&+8.20&-1.63&+2.50\\
    \hdashline
    \rowcolor{gray!10}\textbf{(w/ Omni-CoT)}&\textbf{70.03}&\textbf{76.15}&\textbf{75.66}&\textbf{81.00}&\textbf{64.53}&\textbf{78.51}&\textbf{82.60}&\textbf{42.98}&\textbf{55.20}&\textbf{71.02}&\textbf{62.00}\\
    \rowcolor{gray!10}$\Delta(\uparrow)$&+7.41&+0.33&+1.65&+6.00&+6.97&+0.82&+26.40&+3.36&+8.60&+0.82&+2.5\\
    \midrule
    \textbf{GPT-4o}&55.79&74.43&67.76&65.50&49.42&74.38&43.24&32.49&39.60&57.55&53.50\\
    \hdashline
    (w/ Viewpoint Guiding)&\textbf{61.67}&73.69&65.28&\textbf{74.00}&\textbf{54.07}&\textbf{76.03}&\textbf{70.06}&\textbf{37.31}&\textbf{39.80}&53.88&\textbf{56.50}\\
    $\Delta(\uparrow)$&+5.88&-0.74&-2.48&+8.50&{+4.65}&{+1.65}&{+26.82}&{+4.82}&{+0.20}&-3.67&{+3.00}\\
    \hdashline
    \rowcolor{gray!10}\textbf{(w/ Omni-CoT)}&\textbf{62.08}&73.77&\textbf{68.42}&\textbf{75.00}&\textbf{54.07}&\textbf{76.03}&\textbf{71.53}&\textbf{37.94}&37.60&52.65&\textbf{58.50}\\
    \rowcolor{gray!10}$\Delta(\uparrow)$&+6.17&-0.66&{+0.66}&+9.50&{+4.65}&+1.65&+28.29&+5.45&-2.00&-4.90&+5.00\\
    \midrule
    \textbf{Gemini-2.0-flash}&{57.12}&73.03&69.41&66.50&52.33&{80.16}&{48.10}&{32.91}&{40.20}&{61.13}&{54.00}\\
    \hdashline
    (w/ Viewpoint Guiding)&\textbf{62.95}&\textbf{73.60}&68.75&\textbf{75.50}&\textbf{55.23}&\textbf{83.47}&\textbf{72.64}&\textbf{35.85}&\textbf{41.00}&\textbf{62.45}&\textbf{54.00}\\
    $\Delta(\uparrow)$&+5.83&+0.57&-0.66&+9.00&+2.90&+3.31&+24.54&+2.94&+0.8&+1.32&+0.00\\
     \hdashline
    \rowcolor{gray!10}\textbf{(w/ Omni-CoT)}&\textbf{63.89}&\textbf{73.77}&\textbf{69.41}&\textbf{76.50}&\textbf{57.56}&\textbf{84.30}&\textbf{74.36}&\textbf{36.06}&\textbf{42.20}&\textbf{62.45}&\textbf{55.50}\\
    \rowcolor{gray!10}$\Delta(\uparrow)$&+6.77&+0.74&+0.00&+10.00&+5.23&+4.14&+26.26&+3.15&+2.00&+1.32&+1.50\\
    \midrule
    \textbf{Qwen2.5-VL-72B}&56.91&{77.38}&68.75&52.00&{58.14}&75.21&{46.75}&32.08&{38.40}&{60.00}&50.00\\
     \hdashline
    (w/ Viewpoint Guiding)&\textbf{64.51}&76.39&\textbf{73.02}&\textbf{66.50}&55.49&\textbf{77.87}&\textbf{76.07}&\textbf{37.31}&\textbf{46.61}&54.89&\textbf{51.00}\\
    $\Delta(\uparrow)$&+7.60&-0.99&+4.27&+14.50&-2.65&+2.66&+29.32&+5.23&+8.21&-5.11&+1.00\\
     \hdashline
    \rowcolor{gray!10}\textbf{(w/ Omni-CoT)}&\textbf{65.41}&76.80&\textbf{74.01}&\textbf{66.50}&54.32&\textbf{77.87}&\textbf{80.12}&\textbf{37.74}&\textbf{45.20}&56.33&\textbf{51.50}\\
    \rowcolor{gray!10}$\Delta(\uparrow)$&+8.50&-0.58&+5.26&+14.50&-3.82&+2.66&+33.37&+5.66&+6.80&-3.67&+1.50\\
    \midrule
    \textbf{InternVL2.5-8B}&52.76&68.52&70.07&60.00&51.74&66.12&45.23&31.24&33.00&44.08&58.00\\
     \hdashline
    (w/ Viewpoint Guiding)&\textbf{55.76}&\textbf{68.77}&\textbf{72.69}&\textbf{63.00}&49.42&\textbf{80.17}&\textbf{52.39}&\textbf{34.38}&\textbf{34.40}&\textbf{48.98}&\textbf{60.50}\\
    $\Delta(\uparrow)$&+3.00&+0.25&+2.62&+3.00&-2.32&+14.05&+7.16&+3.14&+1.40&+4.90&+2.5\\
     \hdashline
    \rowcolor{gray!10}\textbf{(w/ Omni-CoT)}&\textbf{58.04}&\textbf{71.48}&\textbf{72.69}&\textbf{63.50}&\textbf{52.33}&\textbf{80.99}&\textbf{58.28}&\textbf{35.64}&\textbf{34.40}&\textbf{49.39}&\textbf{61.50}\\
    \rowcolor{gray!10}$\Delta(\uparrow)$&+5.28&+2.96&+2.62&+3.50&+0.59&+13.97&+13.05&+4.40&+1.40&+5.31&+3.50\\
    \bottomrule[1pt]
  \end{tabular}}
  \label{tab4-exp}
  \vspace{-5mm}
\end{table}

%% file: tabs/ablation_new.tex

\begin{table}[!t]
\footnotesize
  \centering
  \caption{Ablation studies of Omni-CoT on ODI-Bench}
  \renewcommand\arraystretch{1.2}
  \resizebox{0.8\linewidth}{!}{
  \begin{tabular}{c|ccc|ccc}
    \toprule[1pt]
     Model&\multicolumn{3}{c}{Strategy}&\multicolumn{3}{c}{Performace}\\
     \cmidrule(r){1-1} \cmidrule(r){2-4} \cmidrule(r){5-7}
    &Viewpoint Guiding&Crop Grounding&Crop Refinement&Overall&General&Spatial\\
    \midrule
   {Gemini-2.0-flash}&&&&57.12&70.49&45.05\\
   \hdashline
    &\ding{52}&&&63.07&72.08&54.94\\
    &\ding{52}&\ding{52}&&62.79&71.79&54.67\\
    &\ding{52}&&\ding{52}&58.29&67.67&49.83\\
    &&\ding{52}&\ding{52}&55.88&70.05&43.09\\
    \rowcolor{gray!10} \quad&\ding{52}&\ding{52}&\ding{52}&\textbf{63.89}&\textbf{72.63}&\textbf{56.01}\\
    \midrule
    {InternVL2.5-8B}&&&&52.76&66.33&40.53\\
    \hdashline
    &\ding{52}&&&55.76&67.82&44.88\\
    &\ding{52}&\ding{52}&&53.71&65.79&42.83\\
    &\ding{52}&&\ding{52}&48.93&54.29&44.12\\
    &&\ding{52}&\ding{52}&50.52&66.78&35.85\\
    \rowcolor{gray!10} \quad&\ding{52}&\ding{52}&\ding{52}&\textbf{58.04}&\textbf{69.81}&\textbf{47.43}\\
    \bottomrule[1pt]
  \end{tabular}}
  \vspace{-3mm}
  \label{tabs:ablation}
  \end{table}

%% file: tabs/hyperparameter.tex
\begin{table}[t]
\vspace{-1.5em}
\setlength{\belowcaptionskip}{-0.01cm}
\centering
\belowrulesep=0pt
\aboverulesep=0pt
\renewcommand\arraystretch{1.2}
\caption{More hyperparameter ablation of \textbf{Omni-CoT} on ODI-Bench, best performances are \textbf{bolded}.}
\scriptsize
   \resizebox{\linewidth}{!}{
   \begin{tabular}{l
>{\centering\arraybackslash}p{0.85cm}
>{\centering\arraybackslash}p{0.85cm}
>{\centering\arraybackslash}p{0.85cm}
>{\centering\arraybackslash}p{0.85cm}
>{\centering\arraybackslash}p{0.85cm}
>{\centering\arraybackslash}p{0.85cm}
>{\centering\arraybackslash}p{0.85cm}
>{\centering\arraybackslash}p{0.85cm}
>{\centering\arraybackslash}p{0.85cm}
>{\centering\arraybackslash}p{0.85cm}
>{\centering\arraybackslash}p{0.85cm} 
}
    \toprule[1pt]
     \multirow{2}{*}{\textit{Method}}&\multirow{2}{*}{\textit{Overall}}&\multicolumn{5}{c}{\textit{General}}&\multicolumn{5}{c}{\textit{Spatial}}\\
  \cmidrule(lr){3-7} \cmidrule(lr){8-12}
  &&OA&HA&Exist.&Count.&OCR&EVO&AVO&SS&RD&OR\\
  \midrule
    \textbf{GPT-4o}&55.79&\textbf{74.43}&67.76&65.50&49.42&74.38&43.24&32.49&39.60&\textbf{57.55}&53.50\\
    \hdashline
    Omni-CoT (80$^{\circ}$ FoV)&60.27&66.39&68.09&74.50&52.91&75.21&\textbf{73.37}&37.73&38.60&52.65&58.00\\
    Omni-CoT (100$^{\circ}$ FoV)&60.71 &67.29&67.11&74.50&54.07&73.55&{72.27}&37.73&\textbf{40.40}&{55.10}&\textbf{60.50}\\ 
    \rowcolor{gray!10}\textbf{Omni-CoT (90$^{\circ}$ FoV ) (Ours)}&\textbf{62.08}&73.77&\textbf{68.42}&\textbf{75.00}&\textbf{54.07}&\textbf{76.03}&{71.53}&\textbf{37.94}&37.60&52.65&{58.50}\\
    \bottomrule[1pt]
  \end{tabular}
  \label{tabs:hyperparameter}}
  \vspace{-5mm}
\end{table}


%% file: tabs/table5-multiview.tex
\begin{table}[t]
\setlength{\belowcaptionskip}{-0.01cm}
\centering
\belowrulesep=0pt
\aboverulesep=0pt
\renewcommand\arraystretch{1.2}
\caption{Performance comparison of multi-view input, video input and proposed Omni-CoT, best performances are \textbf{bolded}.}
\scriptsize
   \resizebox{\linewidth}{!}{
   \begin{tabular}{l
>{\centering\arraybackslash}p{0.85cm}
>{\centering\arraybackslash}p{0.85cm}
>{\centering\arraybackslash}p{0.85cm}
>{\centering\arraybackslash}p{0.85cm}
>{\centering\arraybackslash}p{0.85cm}
>{\centering\arraybackslash}p{0.85cm}
>{\centering\arraybackslash}p{0.85cm}
>{\centering\arraybackslash}p{0.85cm}
>{\centering\arraybackslash}p{0.85cm}
>{\centering\arraybackslash}p{0.85cm}
>{\centering\arraybackslash}p{0.85cm} 
}
    \toprule[1pt]
     \multirow{2}{*}{\textit{Method}}&\multirow{2}{*}{\textit{Overall}}&\multicolumn{5}{c}{\textit{General}}&\multicolumn{5}{c}{\textit{Spatial}}\\
  \cmidrule(lr){3-7} \cmidrule(lr){8-12}
  &&OA&HA&Exist.&Count.&OCR&EVO&AVO&SS&RD&OR\\
  \midrule
    \textbf{GPT-4o}&55.79&\textbf{74.43}&67.76&65.50&49.42&74.38&43.24&32.49&\textbf{39.60}&\textbf{57.55}&53.50\\
    \hdashline
    (multi-view input)&{56.01}&74.10&67.11&{71.00}&{53.49}&{76.03}&{45.15}&{32.91}&37.60&52.24&{54.00}\\
    \rowcolor{gray!10}\textbf{(w/ Omni-CoT)}&\textbf{62.08}&73.77&\textbf{68.42}&\textbf{75.00}&\textbf{54.07}&\textbf{76.03}&\textbf{71.53}&\textbf{37.94}&37.60&52.65&\textbf{58.50}\\
    \midrule
    \textbf{Gemini-2.0-flash}&{57.12}&73.03&69.41&66.50&52.33&{80.16}&{48.10}&{32.91}&{40.20}&{61.13}&{54.00}\\
    \hdashline
    (multi-view input)&54.63&72.95&64.46&46.50&55.81&63.64&49.57&28.30&37.00&55.92&55.50\\
    (video input)&55.34&70.66&67.11&67.00&48.84&67.77&48.59&30.40&36.00&\textbf{66.12}&52.50\\
    \rowcolor{gray!10}\textbf{(w/ Omni-CoT)}&\textbf{63.89}&\textbf{73.77}&\textbf{69.41}&\textbf{76.50}&\textbf{57.56}&\textbf{84.30}&\textbf{74.36}&\textbf{36.06}&\textbf{42.20}&{62.45}&\textbf{55.50}\\
    \midrule
    \textbf{InternVL2.5-8B}&52.76&68.52&70.07&60.00&51.74&66.12&45.23&31.24&33.00&44.08&58.00\\
     \hdashline
    (multi-view input)&51.97&70.49&71.38&62.50&47.67&76.86&39.14&22.22&33.60&51.02&58.00\\
    (video input)&50.63&68.85&71.38&\textbf{64.00}&43.60&61.16&36.32&23.27&34.00&\textbf{53.88}&60.00\\
    \rowcolor{gray!10}\textbf{(w/ Omni-CoT)}&\textbf{58.04}&\textbf{71.48}&\textbf{72.69}&{63.50}&\textbf{52.33}&\textbf{80.99}&\textbf{58.28}&\textbf{35.64}&\textbf{34.40}&{49.39}&\textbf{61.50}\\
    \bottomrule[1pt]
  \end{tabular}
  \label{tab5-multiview}}
  \vspace{-5mm}
\end{table}


%% file: figures/odi.tex
\begin{figure}[t]
\begin{center}
\includegraphics[width=0.9\linewidth]{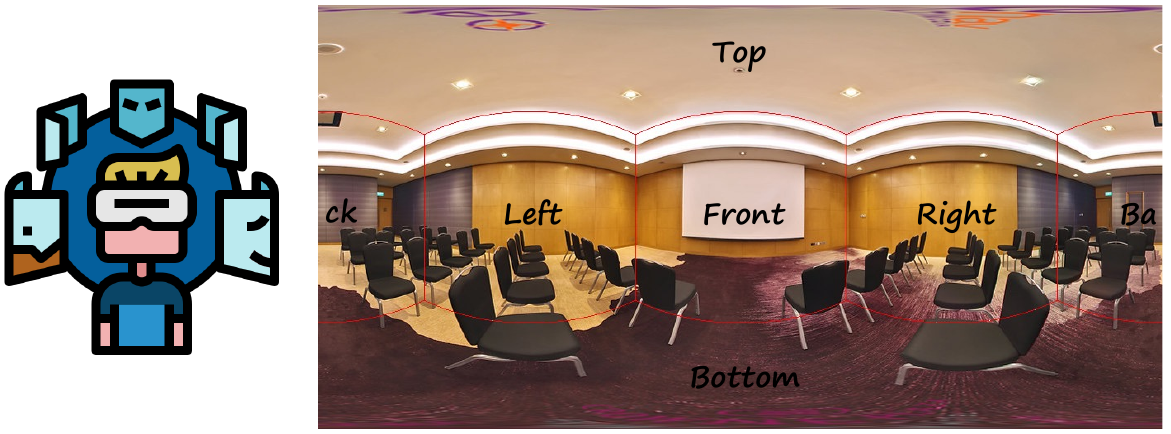}
\end{center}
\vspace{-1.5em}
\caption{Illustration of omnidirectional image browsing. The ODI is viewed using a VR head-mounted display, with the corresponding viewpoints on the ERP projection shown in the right panel.}
\label{figs:odi}
\vspace{-1em}
\end{figure}

%% file: figures/all_tasks.tex
\begin{figure}[t]
\begin{center}
\includegraphics[width=\linewidth]{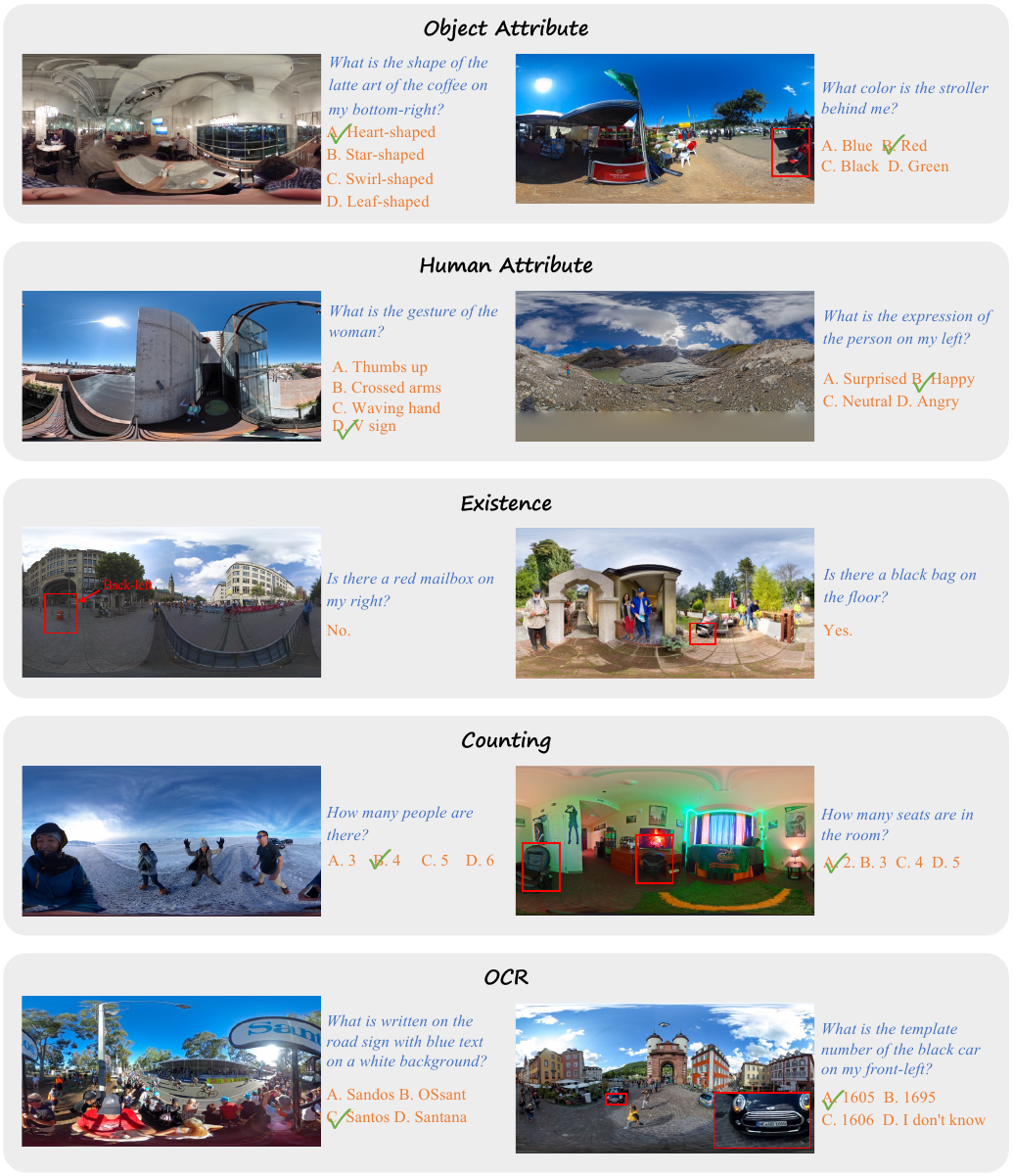}
\end{center}
\vspace{-1em}
\caption{Examples of general-level tasks in ODI-Bench.}
\label{figs:all_tasks}
\vspace{-2mm}
\end{figure}

%% file: figures/all_tasks_spatial.tex
\begin{figure}[t]
\begin{center}
\includegraphics[width=\linewidth]{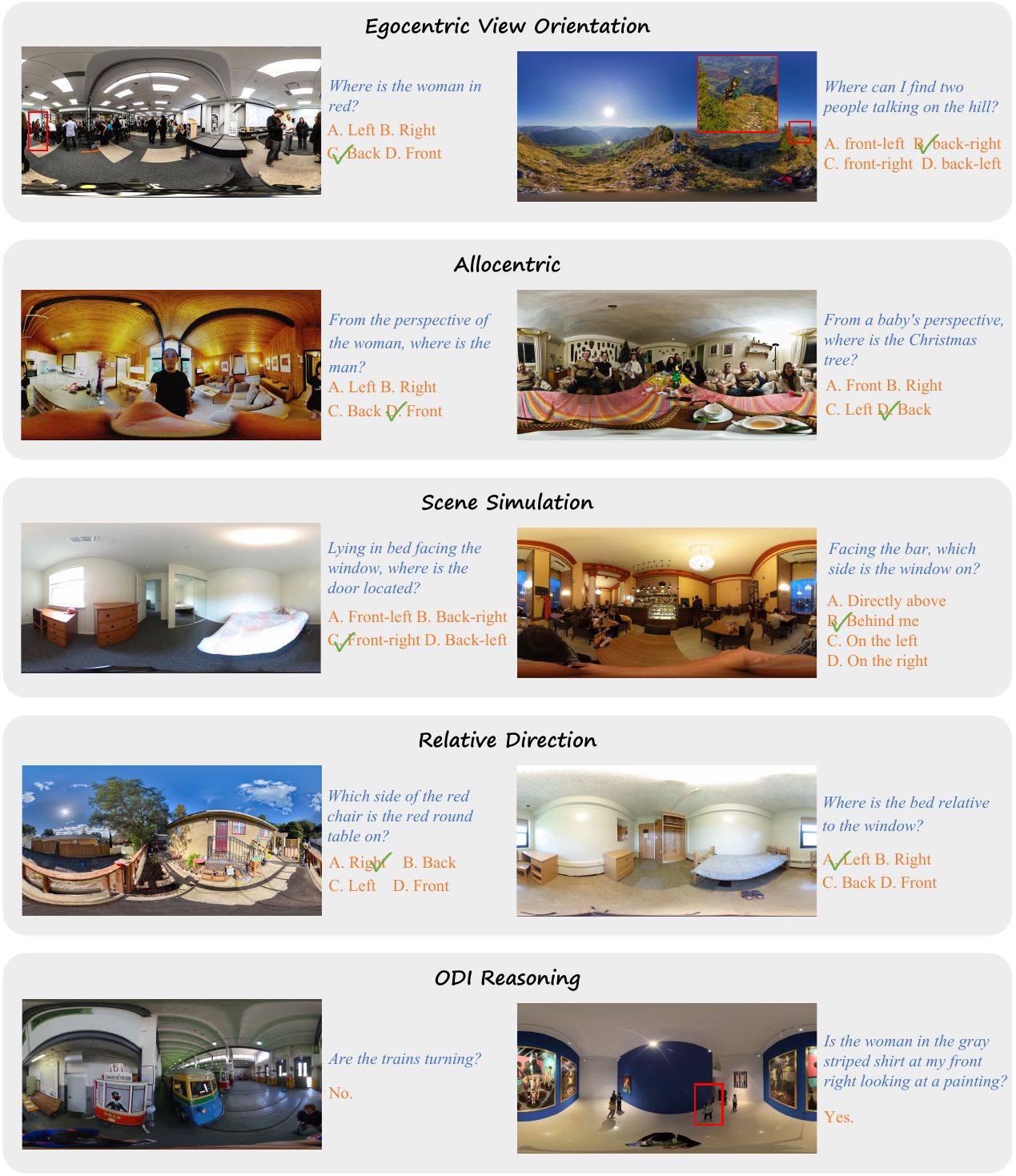}
\end{center}
\vspace{-1em}
\caption{Examples of spatial-level tasks in ODI-Bench.}
\label{figs:all_tasks_spatial}
\vspace{-2mm}
\end{figure}

%% file: figures/wordcloud.tex
\begin{wrapfigure}{r}{0.4\textwidth} 
\vspace{-1em}
\begin{center}
\includegraphics[width=\linewidth]{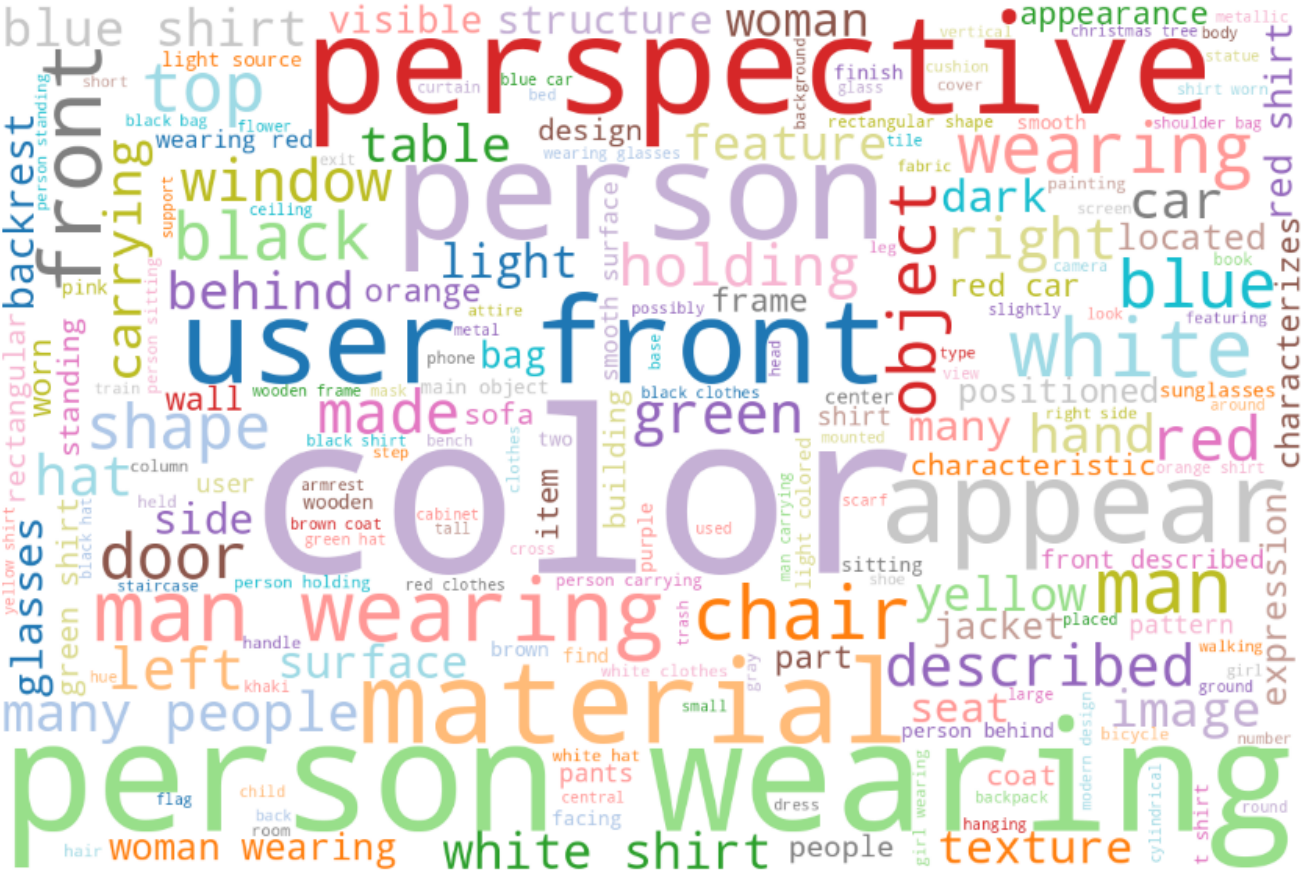}
\end{center}
\vspace{-1em}
\caption{Wordcloud of ODI-Bench.}
\label{figs:wordcloud}
\vspace{-2mm}
\end{wrapfigure}
 

%% file: figures/error_analysis.tex
\begin{figure}[t]
\vspace{-2.5em}
\begin{center}
\includegraphics[width=1\linewidth]{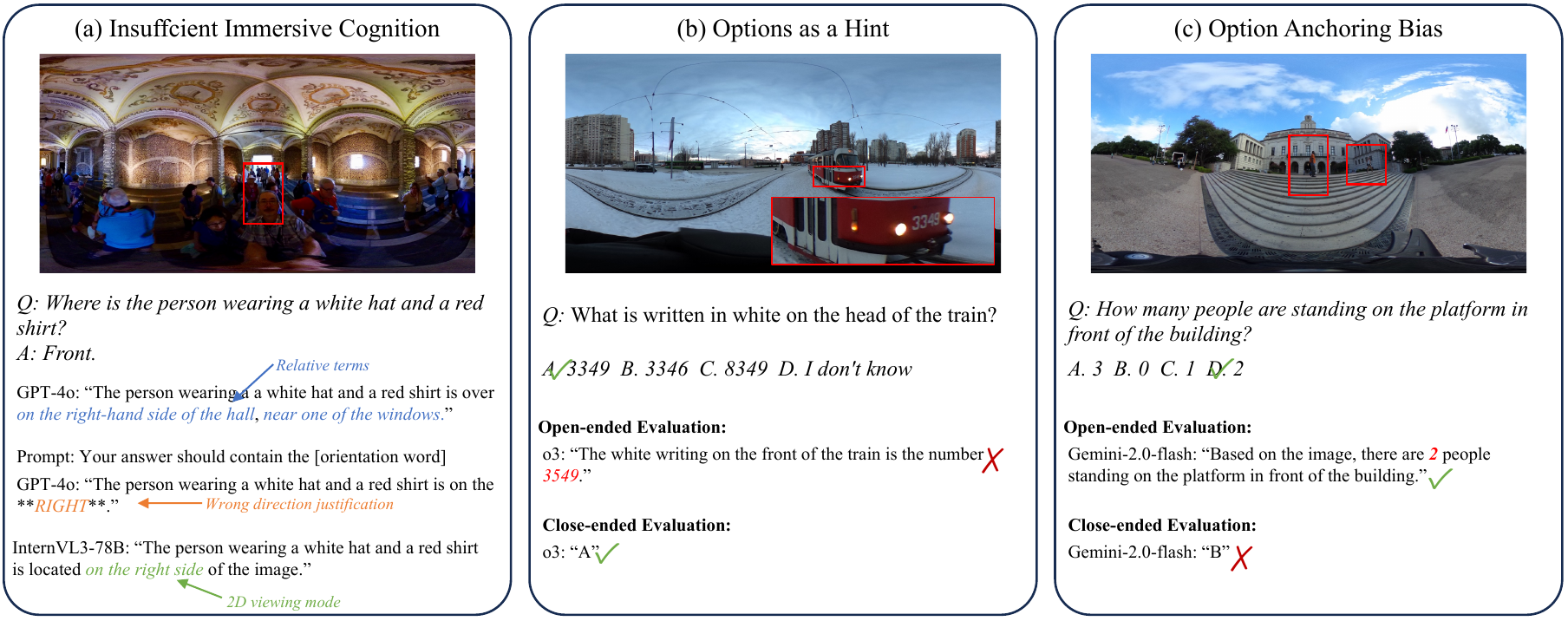}
\end{center}
\vspace{-1em}
\caption{Different error cases in the ODI-Bench.}
\label{figs:error_case}
\end{figure}

%% file: figures/zero_shot_omni_cot.tex
\begin{figure}[t]
\begin{center}
\includegraphics[width=1\linewidth]{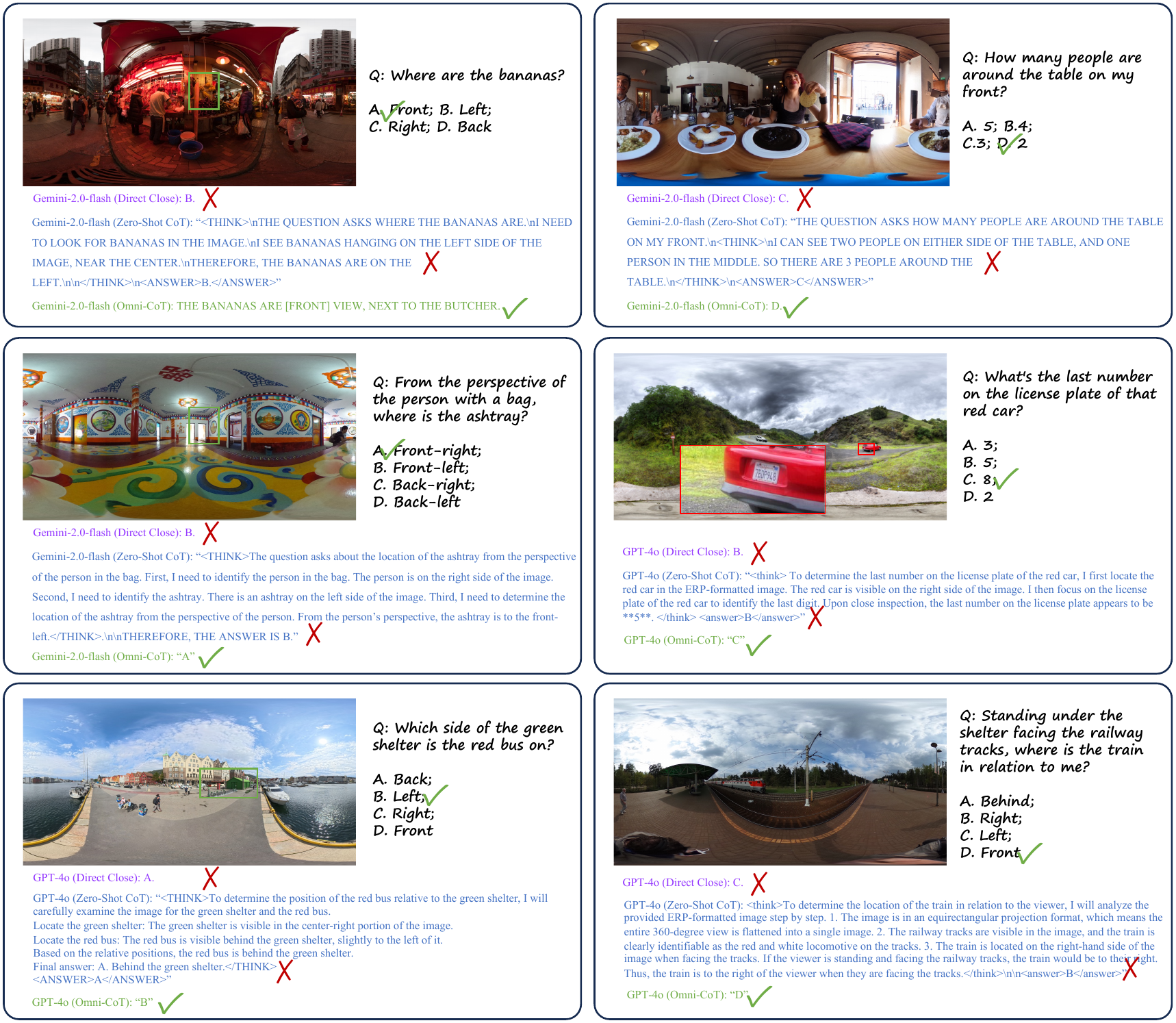}
\end{center}
\vspace{-1em}
\caption{Qualitative examples of direct answering, Zero-shot CoT and proposed Omni-CoT.}
\label{figs:zero_shot_omni_cot}
\vspace{-5mm}
\end{figure}

%% file: tabs/zero-shot-cot.tex
\begin{table}
\setlength{\belowcaptionskip}{-0.01cm}
\centering
\belowrulesep=0pt
\aboverulesep=0pt
\renewcommand\arraystretch{1.2}
\caption{Performance of Zero-shot CoT and Omni-CoT on ODI-Bench, better performances over baseline are \textbf{bolded}.}
\scriptsize
   \resizebox{\linewidth}{!}{
   \begin{tabular}{l
>{\centering\arraybackslash}p{0.85cm}
>{\centering\arraybackslash}p{0.85cm}
>{\centering\arraybackslash}p{0.85cm}
>{\centering\arraybackslash}p{0.85cm}
>{\centering\arraybackslash}p{0.85cm}
>{\centering\arraybackslash}p{0.85cm}
>{\centering\arraybackslash}p{0.85cm}
>{\centering\arraybackslash}p{0.85cm}
>{\centering\arraybackslash}p{0.85cm}
>{\centering\arraybackslash}p{0.85cm}
>{\centering\arraybackslash}p{0.85cm} 
}
    \toprule[1pt]
     \multirow{2}{*}{\textit{Method}}&\multirow{2}{*}{\textit{Overall}}&\multicolumn{5}{c}{\textit{General}}&\multicolumn{5}{c}{\textit{Spatial}}\\
  \cmidrule(lr){3-7} \cmidrule(lr){8-12}
  &&OA&HA&Exist.&Count.&OCR&EVO&AVO&SS&RD&OR\\
    \midrule
    {GPT-4o}&55.79&74.43&67.76&65.50&49.42&74.38&43.24&32.49&39.60&57.55&53.50\\
    \hdashline
    \rowcolor{gray!10}(w/ Zero-shot CoT)&56.18&73.11&66.12&69.00&\textbf{54.07}&75.21&43.80&37.32&39.40&56.73&52.00\\
    \rowcolor{gray!10}$\Delta(\uparrow)$&+0.39&-1.32&-1.64&+3.50&+4.65&+0.83&+0.56&+4.83&-0.20&-0.82&-1.50\\
     \hdashline
    \rowcolor{gray!10}(w/ Omni-CoT)&\textbf{62.08}&\textbf{73.77}&\textbf{68.42}&\textbf{75.00}&\textbf{54.07}&\textbf{76.03}&\textbf{71.53}&\textbf{37.94}&37.60&52.65&\textbf{58.50}\\
    \rowcolor{gray!10}$\Delta(\uparrow)$&+6.17&-0.66&{+0.66}&+9.50&{+4.65}&+1.65&+28.29&+5.45&-2.00&-4.90&+5.00\\
    \midrule
    {Gemini-2.0-flash}&{57.12}&73.03&69.41&66.50&52.33&{80.16}&{48.10}&{32.91}&{40.20}&{61.13}&{54.00}\\
     \hdashline
    \rowcolor{gray!10}(w/ Zero-shot CoT)&57.29&72.70&69.08&70.50&48.26&82.64&49.20&31.87&40.80&61.22&54.50\\
    \rowcolor{gray!10}$\Delta(\uparrow)$&+0.17&-0.33&-0.33&+4.00&-4.07&+2.48&+1.10&-1.04&+0.60&+0.09&+0.50\\
     \hdashline
    \rowcolor{gray!10}(w/ Omni-CoT)&\textbf{63.89}&\textbf{73.77}&{69.41}&\textbf{76.50}&\textbf{57.56}&\textbf{84.30}&\textbf{74.36}&\textbf{36.06}&\textbf{42.20}&\textbf{62.45}&\textbf{55.50}\\
    \rowcolor{gray!10}$\Delta(\uparrow)$&+6.77&+0.74&+0.00&+10.00&+5.23&+4.14&+26.26&+3.15&+2.00&+1.32&+1.50\\
    \bottomrule[1pt]
  \end{tabular}}
  \label{tabs:zero-shot-cot}
\end{table}

%% file: tabs/openended-omni-cot.tex
\begin{table}[t]
\setlength{\belowcaptionskip}{-0.01cm}
\centering
\belowrulesep=0pt
\aboverulesep=0pt
\renewcommand\arraystretch{1.2}
\caption{Performance of Omni-CoT on ODI-Bench under the open-ended evaluation setting, better performances over baseline are \textbf{bolded}.}
\scriptsize
   \resizebox{\linewidth}{!}{
   \begin{tabular}{l
>{\centering\arraybackslash}p{0.85cm}
>{\centering\arraybackslash}p{0.85cm}
>{\centering\arraybackslash}p{0.85cm}
>{\centering\arraybackslash}p{0.85cm}
>{\centering\arraybackslash}p{0.85cm}
>{\centering\arraybackslash}p{0.85cm}
>{\centering\arraybackslash}p{0.85cm}
>{\centering\arraybackslash}p{0.85cm}
>{\centering\arraybackslash}p{0.85cm}
>{\centering\arraybackslash}p{0.85cm}
>{\centering\arraybackslash}p{0.85cm} 
}
    \toprule[1pt]
     \multirow{2}{*}{\textit{Method}}&\multirow{2}{*}{\textit{Overall}}&\multicolumn{5}{c}{\textit{General}}&\multicolumn{5}{c}{\textit{Spatial}}\\
  \cmidrule(lr){3-7} \cmidrule(lr){8-12}
  &&OA&HA&Exist.&Count.&OCR&EVO&AVO&SS&RD&OR\\
    \midrule
    InternVL2.5-8B&30.86&33.52&22.34&62.00&39.53&34.71&21.96&25.68&20.40&38.98&51.55\\
     \hdashline
    \rowcolor{gray!10}(w/ Omni-CoT)&\textbf{36.13}&\textbf{37.71}&\textbf{30.46}&\textbf{67.50}&\textbf{39.53}&\textbf{41.74}&\textbf{33.50}&\textbf{27.46}&\textbf{23.40}&\textbf{42.04}&\textbf{53.50}\\
    \rowcolor{gray!10}$\Delta(\uparrow)$&+5.27&+4.19&{+8.12}&+5.50&{+0.00}&+7.03&+11.54&+1.78&+3.00&+3.06&+1.95\\
    \midrule
    Gemini-2.0-flash&36.42&37.82&28.55&48.26&{50.00}&{56.50}&30.86&25.89&\underline{31.00}&55.51&42.17\\
     \hdashline
    \rowcolor{gray!10}(w/ Omni-CoT)&\textbf{45.20}&\textbf{38.51}&\textbf{29.41}&\textbf{68.50}&\textbf{52.33}&\textbf{58.26}&\textbf{64.00}&\textbf{33.54}&{30.70}&\textbf{58.16}&\textbf{44.20}\\
    \rowcolor{gray!10}$\Delta(\uparrow)$&+8.78&+0.69&+0.86&+20.24&+2.33&+1.76&+33.14&+7.65&-0.30&+2.65&+2.03\\
    \bottomrule[1pt]
  \end{tabular}}
  \label{tabs:omni-cot-openeded}
\end{table}

%% file: tabs/Filter_ratio.tex
\begin{table}
\vspace{-1em}
\setlength{\belowcaptionskip}{-0.01cm}
\centering
\belowrulesep=0pt
\aboverulesep=0pt
\renewcommand\arraystretch{1.2}
\caption{Filter ratio of Crop Refinement Step on GPT-4o.}
\scriptsize
   \resizebox{\linewidth}{!}{
   \begin{tabular}{l
>{\centering\arraybackslash}p{0.85cm}
>{\centering\arraybackslash}p{0.85cm}
>{\centering\arraybackslash}p{0.85cm}
>{\centering\arraybackslash}p{0.85cm}
>{\centering\arraybackslash}p{0.85cm}
>{\centering\arraybackslash}p{0.85cm}
>{\centering\arraybackslash}p{0.85cm}
>{\centering\arraybackslash}p{0.85cm}
>{\centering\arraybackslash}p{0.85cm}
>{\centering\arraybackslash}p{0.85cm} 
}
    \toprule[1pt]
    \multicolumn{5}{c}{\textit{General}}&\multicolumn{5}{c}{\textit{Spatial}}\\
  \cmidrule(lr){1-5} \cmidrule(lr){6-10}
  OA&HA&Exist.&Count.&OCR&EVO&AVO&SS&RD&OR\\
    \midrule
    35.60\%&45.83\%&63.96\%&49.59\%&55.56\%&54.44\%&45.71\%&40.50\%&55.90\%&45.71\%\\
    \bottomrule[1pt]
  \end{tabular}}
  \label{tabs:Filter_ratio}
\end{table}


%% file: tabs/resize.tex
\begin{table}[t]
\setlength{\belowcaptionskip}{-0.01cm}
\centering
\belowrulesep=0pt
\aboverulesep=0pt
\renewcommand\arraystretch{1.2}
\caption{Performance of simple resizing on ODI-Bench.}
\scriptsize
   \resizebox{\linewidth}{!}{
   \begin{tabular}{l
>{\centering\arraybackslash}p{0.85cm}
>{\centering\arraybackslash}p{0.85cm}
>{\centering\arraybackslash}p{0.85cm}
>{\centering\arraybackslash}p{0.85cm}
>{\centering\arraybackslash}p{0.85cm}
>{\centering\arraybackslash}p{0.85cm}
>{\centering\arraybackslash}p{0.85cm}
>{\centering\arraybackslash}p{0.85cm}
>{\centering\arraybackslash}p{0.85cm}
>{\centering\arraybackslash}p{0.85cm}
>{\centering\arraybackslash}p{0.85cm} 
}
    \toprule[1pt]
     \multirow{2}{*}{\textit{Method}}&\multirow{2}{*}{\textit{Overall}}&\multicolumn{5}{c}{\textit{General}}&\multicolumn{5}{c}{\textit{Spatial}}\\
  \cmidrule(lr){3-7} \cmidrule(lr){8-12}
  &&OA&HA&Exist.&Count.&OCR&EVO&AVO&SS&RD&OR\\
    \midrule
    {GPT-4o}&55.79&74.43&67.76&65.50&49.42&74.38&43.24&32.49&39.60&57.55&53.50\\
    \hdashline
    (w/ simple resizing)&54.08&72.70&66.45&60.50&48.83&49.59&43.80&37.32&33.12&56.32&54.00\\
    \midrule
    {InternVL3-78B}&{59.43}&{79.18}&{77.30}&66.50&{59.30}&{80.99}&46.01&31.67&{40.40}&{60.82}&{58.50}\\
     \hdashline
    (w/ simple resizing)&55.52&74.43&69.08&60.50&{51.16}&59.50&42.58&31.67&39.80&60.41&59.00\\
    \bottomrule[1pt]
  \end{tabular}}
  \label{tabs:resize}
\end{table}

%% file: tabs/multiple_runs.tex
\begin{table}[t]
\setlength{\belowcaptionskip}{-0.01cm}
\centering
\belowrulesep=0pt
\aboverulesep=0pt
\renewcommand\arraystretch{1.2}
\caption{Performance over multiple runs.}
\scriptsize
   \resizebox{\linewidth}{!}{
   \begin{tabular}{l
>{\centering\arraybackslash}p{0.85cm}
>{\centering\arraybackslash}p{0.85cm}
>{\centering\arraybackslash}p{0.85cm}
>{\centering\arraybackslash}p{0.85cm}
>{\centering\arraybackslash}p{0.85cm}
>{\centering\arraybackslash}p{0.85cm}
>{\centering\arraybackslash}p{0.85cm}
>{\centering\arraybackslash}p{0.85cm}
>{\centering\arraybackslash}p{0.85cm}
>{\centering\arraybackslash}p{0.85cm}
>{\centering\arraybackslash}p{0.85cm} 
}
    \toprule[1pt]
     \multirow{2}{*}{\textit{Method}}&\multirow{2}{*}{\textit{Overall}}&\multicolumn{5}{c}{\textit{General}}&\multicolumn{5}{c}{\textit{Spatial}}\\
  \cmidrule(lr){3-7} \cmidrule(lr){8-12}
  &&OA&HA&Exist.&Count.&OCR&EVO&AVO&SS&RD&OR\\
    \midrule
    {GPT-4o (round1)}&55.79&74.43&67.76&65.50&49.42&74.38&43.24&32.49&39.60&57.55&53.50\\
    {GPT-4o (round2)}&55.76&75.32&67.11&66.00&49.42&74.38&42.33&33.12&39.00&55.51&54.00\\
    {GPT-4o (round3)}&55.71&74.43&67.76&65.50&48.84&75.21&42.70&33.96&38.80&56.73&53.50\\
    \hdashline
    {GPT-4o (mean)}&55.75&74.73&67.52&65.67&49.23&74.66&42.76&33.18&39.13&56.59&53.67\\
    {GPT-4o (std)}&0.04&0.51&0.37&0.29&0.33&0.48&0.46&0.74&0.42&1.02&0.29\\
    \midrule
    InternVL2.5-8B (round1)&52.76&68.52&70.07&60.00&51.74&66.12&45.23&31.24&33.00&44.08&{58.00}\\
    {InternVL2.5-8B (round2)}&{53.19}&{69.02}&{70.39}&60.00&{51.16}&{66.94}&46.01&31.03&{34.20}&{45.71}&{56.50}\\
    {InternVL2.5-8B (round3)}&{53.24}&{69.51}&{71.05}&60.50&{50.00}&{67.77}&46.01&30.40&{33.00}&{46.94}&{56.00}\\
    \hdashline
    InternVL2.5-8B (mean)&53.06&69.02&70.28&60.17&50.96&66.94&45.75&30.89&33.40&45.38&56.83\\
    InternVL2.5-8B (std)&0.26&0.49&0.18&0.28&0.88&0.82&0.45&0.43&0.69&1.44&{1.04}\\
    \bottomrule[1pt]
  \end{tabular}}
  \label{tabs:multiple_runs}
\end{table}


%% file: tabs/prompts.tex
\begin{table}[t]
\vspace{-1em}
\setlength{\belowcaptionskip}{-0.01cm}
\centering
\belowrulesep=0pt
\aboverulesep=0pt
\renewcommand\arraystretch{1.2}
\caption{Comparison of model performance on ODI-Bench under different prompt designs.}
\scriptsize
   \resizebox{\linewidth}{!}{
   \begin{tabular}{l
>{\centering\arraybackslash}p{0.85cm}
>{\centering\arraybackslash}p{0.85cm}
>{\centering\arraybackslash}p{0.85cm}
>{\centering\arraybackslash}p{0.85cm}
>{\centering\arraybackslash}p{0.85cm}
>{\centering\arraybackslash}p{0.85cm}
>{\centering\arraybackslash}p{0.85cm}
>{\centering\arraybackslash}p{0.85cm}
>{\centering\arraybackslash}p{0.85cm}
>{\centering\arraybackslash}p{0.85cm}
>{\centering\arraybackslash}p{0.85cm} 
}
    \toprule[1pt]
     &\multirow{2}{*}{\textit{Overall}}&\multicolumn{5}{c}{\textit{General}}&\multicolumn{5}{c}{\textit{Spatial}}\\
  \cmidrule(lr){3-7} \cmidrule(lr){8-12}
  &&OA&HA&Exist.&Count.&OCR&EVO&AVO&SS&RD&OR\\
    \midrule
    \rowcolor{gray!40}\multicolumn{12}{c}{\itshape Prompt: Answer the question based on the provided ERP-formatted image.}\\
    \midrule
    GPT-4o&55.79&74.43&67.76&65.50&49.42&74.38&43.24&32.49&39.60&57.55&53.50\\
    \midrule
    \rowcolor{gray!40}\multicolumn{12}{c}{\itshape Prompt: This is a 360-degree panoramic image.}\\
    \midrule
    GPT-4o&55.48&74.59&67.43&66.50&48.26&74.38&41.10&32.91&39.60&56.33&55.50\\
    \midrule
    \rowcolor{gray!40}\multicolumn{12}{c}{\itshape Prompt: This is a 360-degree panoramic image, the image is ERP-formatted.}\\
    \midrule
    GPT-4o&56.06&74.92&65.46&66.00&50.58&73.55&43.80&32.91&40.60&56.73&54.00\\
    \bottomrule[1pt]
  \end{tabular}}
  \label{tabs:prompt}
\end{table}

%% file: tabs/inf_time.tex
\begin{table}[t]
\vspace{-1em}
\scriptsize
  \centering
  \caption{Inference time of direct answering, Zero-shot CoT and Omni-CoT.}
  \begin{tabular}{lccccc}
    \toprule[1pt]
     & &{Direct Answering}&Zero-shot CoT&Omni-CoT (only w/ viewpoint guiding)&Full Omni-CoT\\
    \midrule
    {\textbf{InternVL2.5-8B}}&Inference Time&3.44s&7.24s&7.67s&13.81s\\
    \hdashline
    &Overall Performance& 52.76&52.88& 55.76&58.04\\
    \midrule
    \textbf{o3}&Inference Time&12.21s &14.81s &21.11s&35.03s\\
    \hdashline
    &Overall Performance& 62.62&63.89& 68.78&70.03\\
    \bottomrule[1pt]
  \end{tabular}
  \vspace{-3mm}
  \label{tabs:inf_time}
\end{table}

%% file: figures/prompt_cot.tex
\begin{figure}[t]
\begin{center}
\includegraphics[width=1\linewidth]{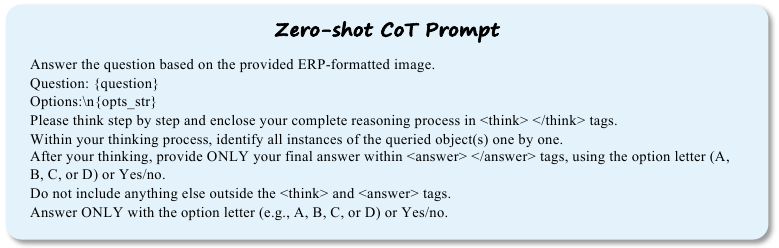}
\end{center}
\vspace{-1.5em}
\caption{Prompts adopted for Zero-shot CoT.}
\label{figs:zero-shot-cot}
\end{figure}

%% file: figures/prompt_exp.tex
\begin{figure}[t]
\begin{center}
\includegraphics[width=1\linewidth]{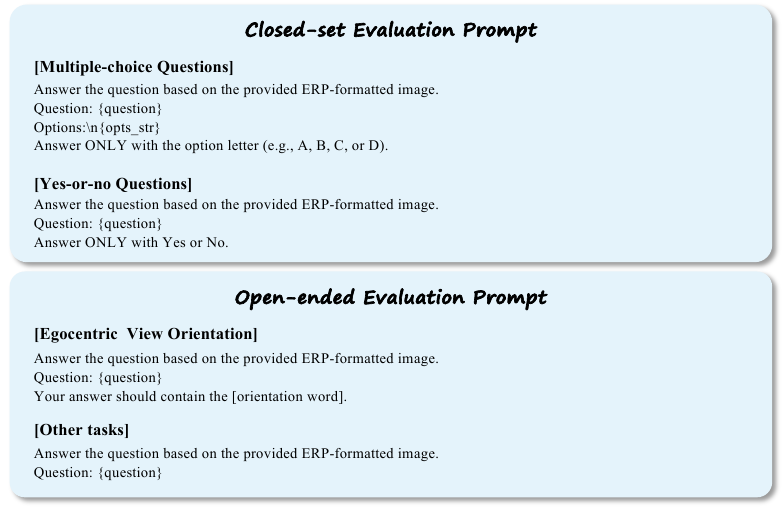}
\end{center}
\vspace{-1.5em}
\caption{Prompts adopted in closed-set and open-ended evaluation experiments.}
\label{figs:prompt_exp}
\vspace{-1em}
\end{figure}

%% file: figures/prompt_qa.tex
\begin{figure}[t]
\begin{center}
\includegraphics[width=1\linewidth]{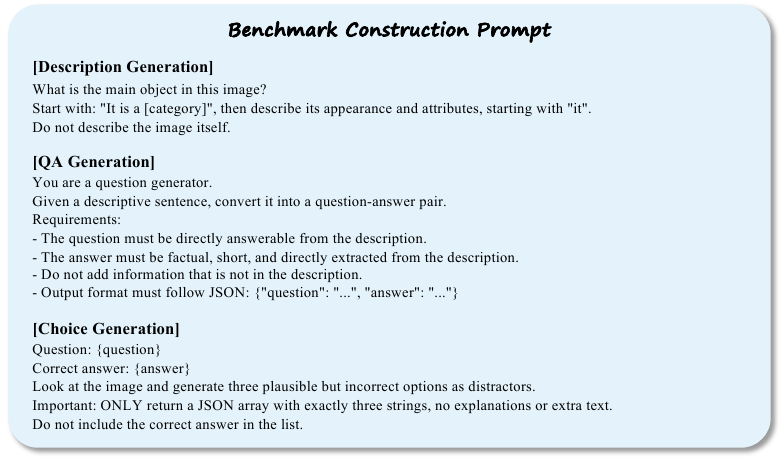}
\end{center}
\vspace{-1.5em}
\caption{Prompts adopted during the benchmark construction process.}
\label{figs:prompt_qa}
\vspace{-1em}
\end{figure}

%% file: figures/prompt_llm.tex
\begin{figure}[t]
\begin{center}
\includegraphics[width=1\linewidth]{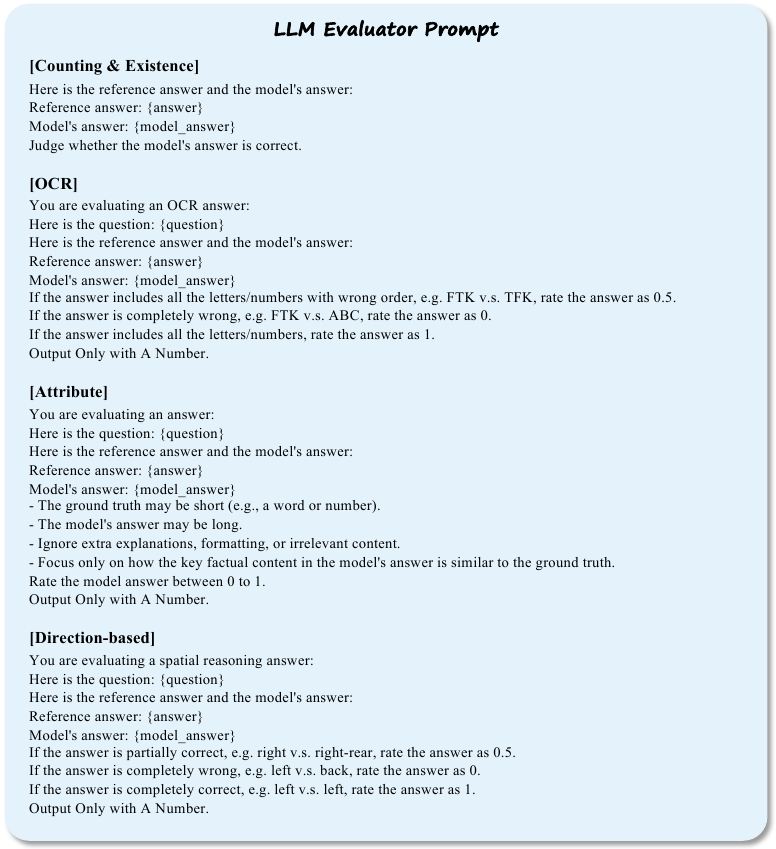}
\end{center}
\vspace{-1.5em}
\caption{Prompts adopted for the LLM-based evaluator.}
\label{figs:prompt_llm}
\vspace{-1em}
\end{figure}

%% file: figures/prompt_omnicot.tex
\begin{figure}[t]
\begin{center}
\includegraphics[width=1\linewidth]{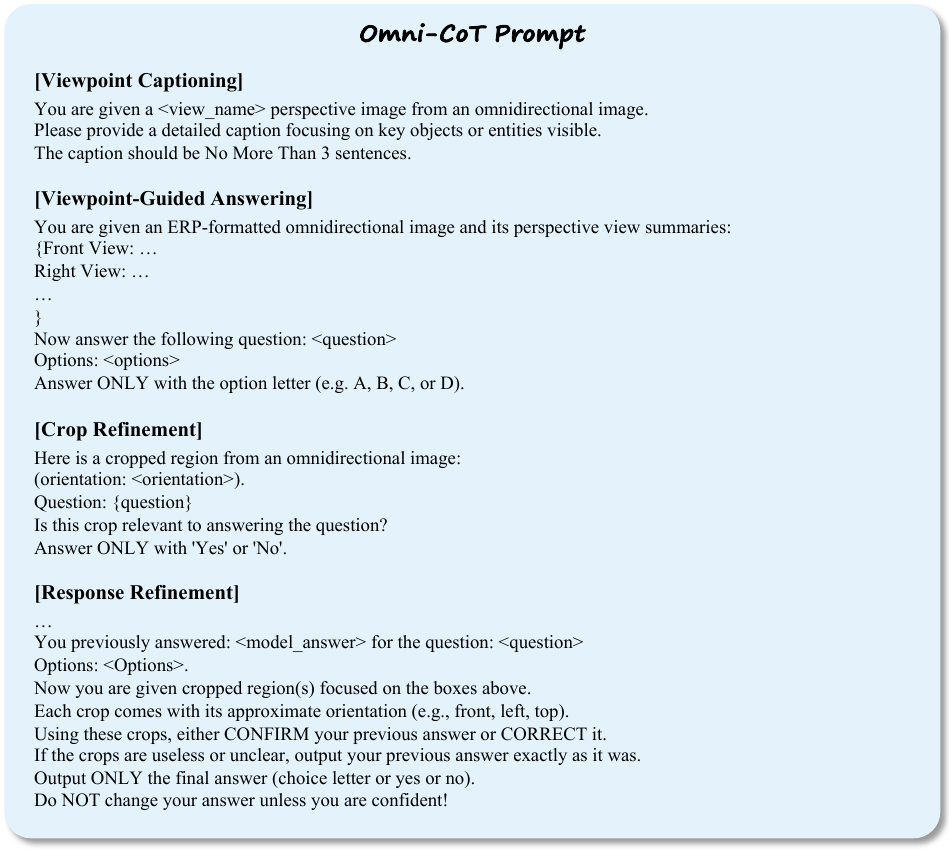}
\end{center}
\vspace{-1.5em}
\caption{Prompts adopted in Omni-CoT.}
\label{figs:prompt_omnicot}
\vspace{-1em}
\end{figure}